\documentclass[letterpaper, 10 pt, conference]{ieeeconf}  

\IEEEoverridecommandlockouts                              

\usepackage{amsmath}
\usepackage{graphicx}
\usepackage{capt-of}
\usepackage{amsfonts}
\usepackage{cleveref}
\usepackage{bbm, dsfont}
\usepackage{booktabs}
\usepackage{multicol}
\usepackage{multirow}
\usepackage{comment}
\usepackage{array}
\usepackage{diagbox}
\usepackage{xcolor}
\usepackage{xspace}

\usepackage{caption}
\makeatletter
\DeclareRobustCommand\onedot{\futurelet\@let@token\@onedot}
\def\@onedot{\ifx\@let@token.\else.\null\fi\xspace}

\makeatother

\title{\LARGE \bf
Breaking the Sub-Millimeter Barrier:\\ Eyeframe Acquisition from Color Images
}

\author{Manel Guzmán$^{1}$ and Antonio Agudo$^{2}$, {\em Member, IEEE}
\thanks{This work has been partially supported by the project GRAVATAR PID2023-151184OB-I00 funded by MCIU/AEI/10.13039/501100011033 and ERDF.}
\thanks{$^{1}$Manel Guzmán is with the Horizons Optical, Sant Cugat del Vallès, Spain.} \thanks{$^{2}$Antonio Agudo is with the Institut de Rob\`otica i Inform\`atica Industrial, CSIC-UPC, Barcelona, Spain.}
}%

\begin{document}

\maketitle

\vspace{-4mm}
\begin{abstract}
Eyeframe lens tracing is an important process in the optical industry that requires sub-millimeter precision to ensure proper lens fitting and optimal vision correction. Traditional frame tracers rely on mechanical tools that need precise positioning and calibration, which are time-consuming and require additional equipment, creating an inefficient workflow for opticians. This work presents a novel approach based on artificial vision that utilizes multi-view information. The proposed algorithm operates on images captured from an InVision system. The full pipeline includes image acquisition, frame segmentation to isolate the eyeframe from background, depth estimation to obtain 3D spatial information, and multi-view processing that integrates segmented RGB images with depth data for precise frame contour measurement. To this end, different configurations and variants are proposed and analyzed on real data, providing competitive measurements from still color images with respect to other solutions, while eliminating the need for specialized tracing equipment and reducing workflow complexity for optical technicians.
\end{abstract}

\section{INTRODUCTION}

In the manufacturing industry, precision and exactitude are paramount. This is particularly evident in the optical sector, which is recognized for its stringent quality standards, where the most demanding applications require accuracy at the millimeter scale. A representative example is the optical tracer, an instrument employed to measure the geometry of eyeglass frames in order to ensure the correct cutting of lenses. By integrating principles of optics and metrology, optical tracers enable the inspection and analysis of complex surfaces with exceptional precision. For many years, conventional tracers have been utilized to capture the geometry of the wide variety of frames available on the market. However, as industrial practices evolve, there is an increasing demand for enhanced efficiency, higher processing speeds, and improved accuracy.

Optical metrology~\cite{c1} commonly relies on image-based measurements, particularly the analysis of fringe patterns, to reconstruct physical quantities of interest. In interferometric techniques, images are produced through the coherent combination of object and reference beams, resulting in characteristic bright–dark fringe distributions. Methods such as classical interferometry~\cite{c2}, holographic interferometry~\cite{c3}, and digital holography~\cite{c4} extract phase information from these patterns, which arise from the superposition of two smooth coherent wavefronts. Photometric stereo uses a set of images with different illumination conditions to jointly infer the object geometry, reflectance, and lighting~\cite{pestevez_icprw22,agudo_icassp23}. Although these approaches differ in the underlying physics of their image formation processes and in the interpretation of fringe-related parameters, they share a common limitation: their performance typically depends on highly controlled experimental conditions, an assumption that is often impractical in real-world scenarios. That limitation underscores a growing need for innovation in optical metrology. In this context, the integration of computer vision methodologies with optical tracing presents a promising and transformative direction. To avoid the projection of patterns, artificial vision can be considered. In this work we present a novel method to measure eyeframes with great precision, using only RGB images that will be exploited to infer segmentations and depth maps that will help the final system.

Segmentation is a very important problem in artificial vision, where many approaches have been presented~\cite{dsa-ilora}. Techniques based on dilated convolutions and pyramid pooling have significantly advanced segmentation performance~\cite{c19}, leading to solutions such as DeepLabV3+~\cite{c20}, which achieve an effective balance between accuracy and computational efficiency. These methods add a decoder on top of the backbone which is normally modeled by ResNet~\cite{c26} variants, MobileNet~\cite{c28} and EfficientNet~\cite{c29}. Transformer-based architectures have also become increasingly relevant, with models such as the Vision Transformer (ViT)~\cite{c21} and SegFormer~\cite{c22} demonstrating strong performance by capturing long-range pixel dependencies. This capability enables more precise boundary delineation and improved contextual understanding. More recent methods include Segment Anything Model (SAM)~\cite{c23} or its more modern version SAM2~\cite{c24} capable of generalizing across various segmentation tasks with good zero-shot performance. 

Obtaining accurate depth maps has long been a central challenge in the field, as depth constitutes a critical cue for both measurement and reconstruction tasks. Depth-map–based approaches typically employ deep learning for depth estimation and subsequently integrate the estimated maps using traditional fusion algorithms. Among learning-based strategies, depth-map methods~\cite{c32,c35} represent the dominant family because they inherit key advantages of classical pipelines~\cite{c36,c37}, effectively decomposing 3D reconstruction into two stages: depth estimation and depth fusion. Supervised multi-view stereo (MVS) algorithms generally follow a standard processing pipeline comprising feature extraction~\cite{c32, c41,c42,c43,c44}, cost-volume construction~\cite{c32,c45,c46,c47}, cost-volume regularization~\cite{c45,c47}, and depth inference. More recently, architectures such as Depth Anything~\cite{c72} have been proposed to predict depth from single images, drawing inspiration from transformer-based models~\cite{c55}. These methods achieve superior zero-shot depth estimation performance compared to previous models such as MiDaS~\cite{c73} and ZoeDepth~\cite{c74}.

In this paper, we present a computer vision-based algorithm for optical frame measurement, examining whether our method can achieve results that meet optical quality standards --it must provide sufficient precision to measure the frame radius for lens edging-- while simplifying and accelerating the measurement process. To frame the problem, the study will concentrate on full-frame glasses, excluding semi-rimless or rimless designs.

\section{METHOD}
\label{section_method}

In this section we introduce the algorithm to extract the final eye-frame trace. We first present the acquisition system we use, the segmentation pipeline to extract the glasses, the way to infer depth maps and finally, the trace measurement system that exploit both segmentation and depth information together with multi-view RGB data. In our final system, we describe different modalities to solve the problem.

\subsection{InVision system}
\begin{figure}[t!]
\centering
\includegraphics[width=8.4cm]{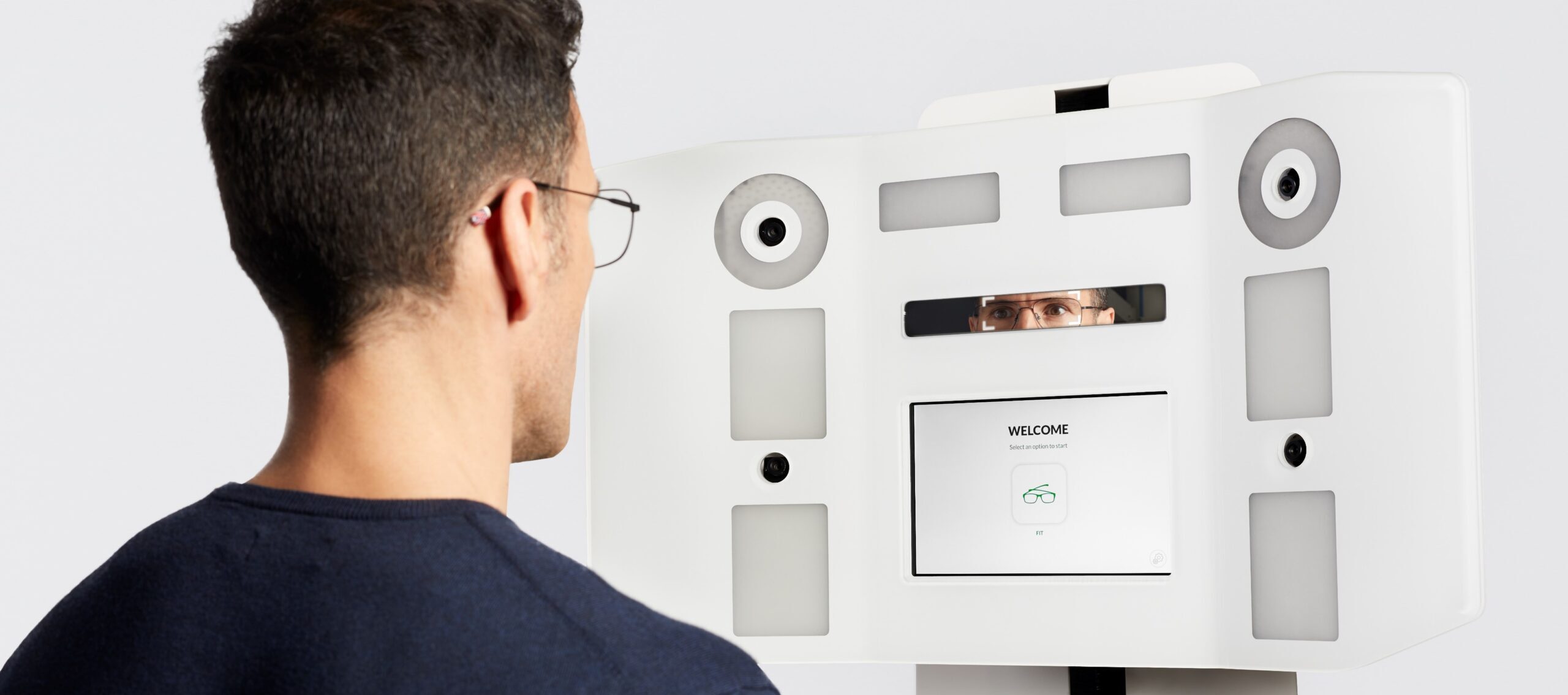}
\caption{\textbf{Invision Head.} The acquisition system consists of four calibrated cameras and two sources of light: visible and IR. Just the 1296$\times$1296 RGB images are considered in this paper.}
\label{fig:invision}
\end{figure}

 For acquisition, we propose using a tower-mounted InVision system (see Fig.~\ref{fig:invision}) that is used in literature as a centering solution for eyeglasses. Using it, we observe users who wear the eyeframe at approximately 50 cm. It has four calibrated cameras and works with two types of lighting: visible and infrared. A single shot captures images from the four cameras synchronously. The pipeline for the cameras does the required color corrections and rectifies the images based on the distortion coefficients of the camera.

\subsection{Segmentation}
\label{segmentation_task}
We now introduce the main details in our segmentation approach, including the dataset we use as well as the corresponding algorithm.

\noindent \textbf{Dataset.} One of the key points in the segmentation step is the creation of a custom dataset for a finetuning process. The dataset is composed of real measures from more than 20 InVision systems distributed around the world. As noted before, the images are taken by groups of four cameras per InVision measure. The final set contains 1,002 images, each one with a labeled mask by exploiting the tool CVAT.ai~\cite{c25}. The dataset includes all possible camera views.

Before proceeding and to enhance the model's ability to generalize and perform robustly on unseen data, we adapt our dataset by means of normalization and data augmentation as follows. First, the pixels are normalized for each channel with a mean and standard deviation. Their values will be computed for each channel independently. This standardization is crucial for neural network training stability and consistency, ensuring that all input features are on a similar scale. Second, a comprehensive data augmentation strategy is implemented to expand and diversify the training set, improving model generalization. The applied transformations include: geometric transformations, noise injection, color and contrast adjustments, blur and sharpness variations, as well as color space transformations. All augmentations are applied stochastically, i.e., governed by predefined probabilities, ensuring that each training epoch presents a unique set of transformed samples. This diversity improves the ability of the model to generalize to real-world and unseen data.

\noindent \textbf{Method.} Inspired by SAM2~\cite{c24}, we present our approach to handle the segmentation problem. Particularly, the model is a transformer architecture with streaming memory for real-time video processing. At the core of SAM2~\cite{c24} is an image encoder that uses a pre-trained Hiera~\cite{c64} model based on the masked autoencoders architecture. This encoder generates feature embeddings for each frame. The memory attention mechanism then conditions the current frame features on past ones, predictions, and new prompts using transformer blocks with self-attention and cross-attention mechanisms. Moreover, this method incorporates a memory system to maintain context across frames. The memory encoder generates memories by combining downsampled output masks with unconditioned frame embeddings. Using hierarchical features during decoding for enhanced performance, it predicts multiple masks and propagates the highest Intersection over Union (IoU) one across the frames. The model also includes a dedicated head to predict whether the object of interest is present in the current frame. To preserve fine details, the method also incorporates skip connections from the hierarchical image encoder to the mask decoder. This model is fine-tuned in our dataset, demonstrating high adaptability across diverse environments and computational resources. Following SAM2~\cite{c24}, we also propose four models denoted by {\em tiny}, {\em small}, {\em base+} and {\em large}, and they vary in the number of parameters of each model. The model is designed to segment the background while preserving the frame structure. This approach significantly reduces the complexity of subsequent processing steps. Small residual regions of the human face appearing around the frame boundaries are not considered a significant limitation for the model's performance.

\subsection{Depth estimation}
A key point in-depth estimation is how it is represented in the model’s output and how it should be fed into the final model for trace prediction. In computer vision, a dense map is a common approach for depth representation. Each pixel in the map corresponds to a value, which indicates the distance between that point in the scene and the camera. The map provides depth information for every pixel in the image, rather than just sparse points as LIDAR does~\cite{perez_itsc24}. The maps show the distance between the camera and the object at each pixel. There exist two common approaches to represent depth: absolute and relative. For the absolute case, the depth values represent real-world distances, like millimeters or centimeters. This is commonly used for tasks like 3D reconstruction and robotics, where precise scale is necessary. For the relative case, the depth values represent depth differences across the images or a segment of the image and usually look like an image showing values from 0-255, or other common scales. That means the values only show how far objects are from each other rather than from the camera in a real-world metric. This is a common approach in applications like monocular depth estimation, where absolute distances may not be available. They represent closer objects with higher values than those that are farther from the camera. The absolute approach provides real-world distances, although it is highly susceptible to noise and holes in the final results. In contrast, the relative approach does not provide such precise information since the depth values are relative, but it offers a more robust prediction, and therefore, we will use this type of depth map.

\noindent \textbf{Method.} To incorporate depth information into our framework, we rely on depth-map estimation methods previously explored in the literature~\cite{c43,c72,c51}. In our setting, the depth map serves primarily as an auxiliary source of information that supports more accurate predictions within the final trace model. For this purpose, we adopt Depth Anything~\cite{c72}, a pre-trained deep learning model for relative depth estimation that achieves high-quality results without the need for task-specific finetuning and demonstrates strong generalization across diverse scenarios. This characteristic is particularly important for our work, as ground-truth depth measurements are not available.
We consider three model variants differing in encoder capacity: ViT-S (24.8M), ViT-B (97.5M), and ViT-L (335.3M), where the number in parenthesis represents the number of parameters.

\subsection{Trace measurement}
Now, we introduce our trace model by exploiting input RGB images in combination with segmentation mask or depth information that were inferred above.

\noindent \textbf{Dataset.}
A new dataset with 500 measurements (4 images for each one) is introduced for training the frame tracer model that consists of real observations collected using two different \textit{InVision} systems. Each data sample includes the model input and its corresponding frame trace as output. The frame trace is defined as a series of radial measurements, where each one represents the distance (radius) from a defined optical center (the geometric center of the frame) to the inner edge of the frame along a specific angle. Each trace contains 600 points per eye frame, with all data stored in a text file. Frame traces are acquired using a commercial tracer device. Similar to the segmentation task, training a network able to measure the frame trace requires a preprocessing and data augmentation stage, which is even more important when working with a limited dataset.

The preprocessing step consists of \textit{pixel value normalization} applied separately to each image modality, assuring faster training and gradient stability. Furthermore, the labels, i.e., the trace measurements, are normalized using \textit{standard normalization}. 

This normalization ensures that the data have zero mean and unit variance, which improves convergence behavior during network training. For data augmentation, we use the same strategies as in the segmentation task (see section~\ref{segmentation_task}).

\begin{figure}[t!]
    \centering
    \includegraphics[width=0.5\linewidth]{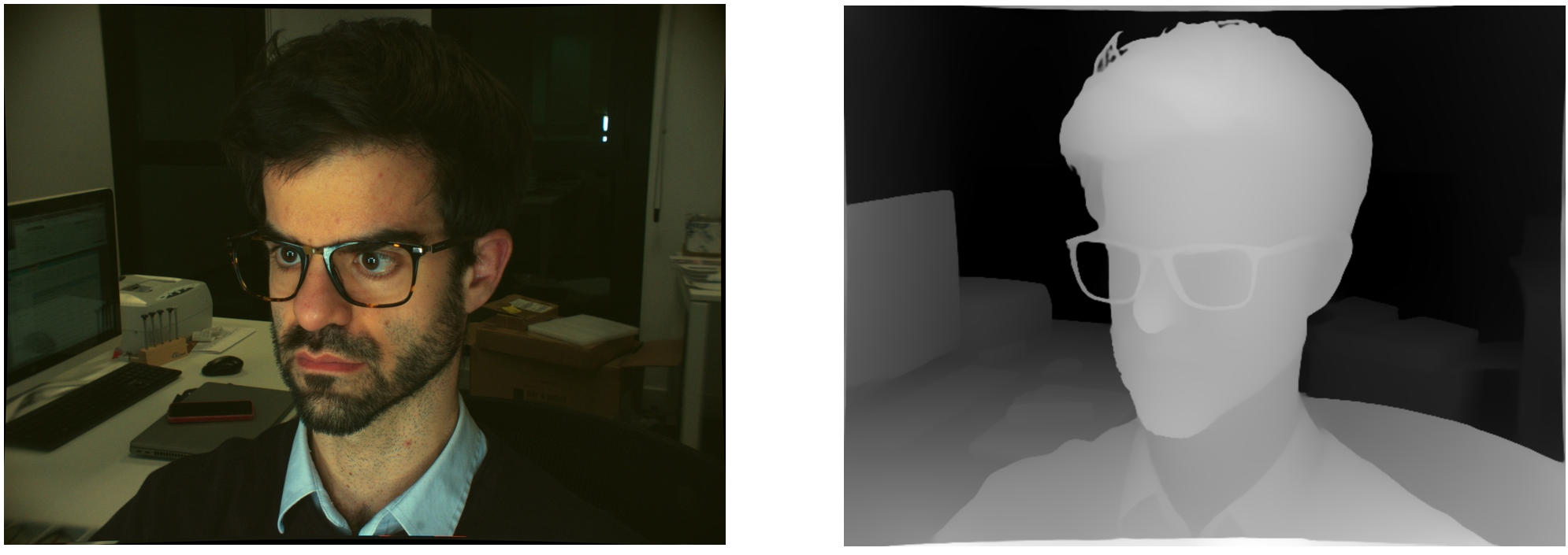}%
    \hspace{0.05\linewidth}%
    \includegraphics[width=0.41\linewidth]{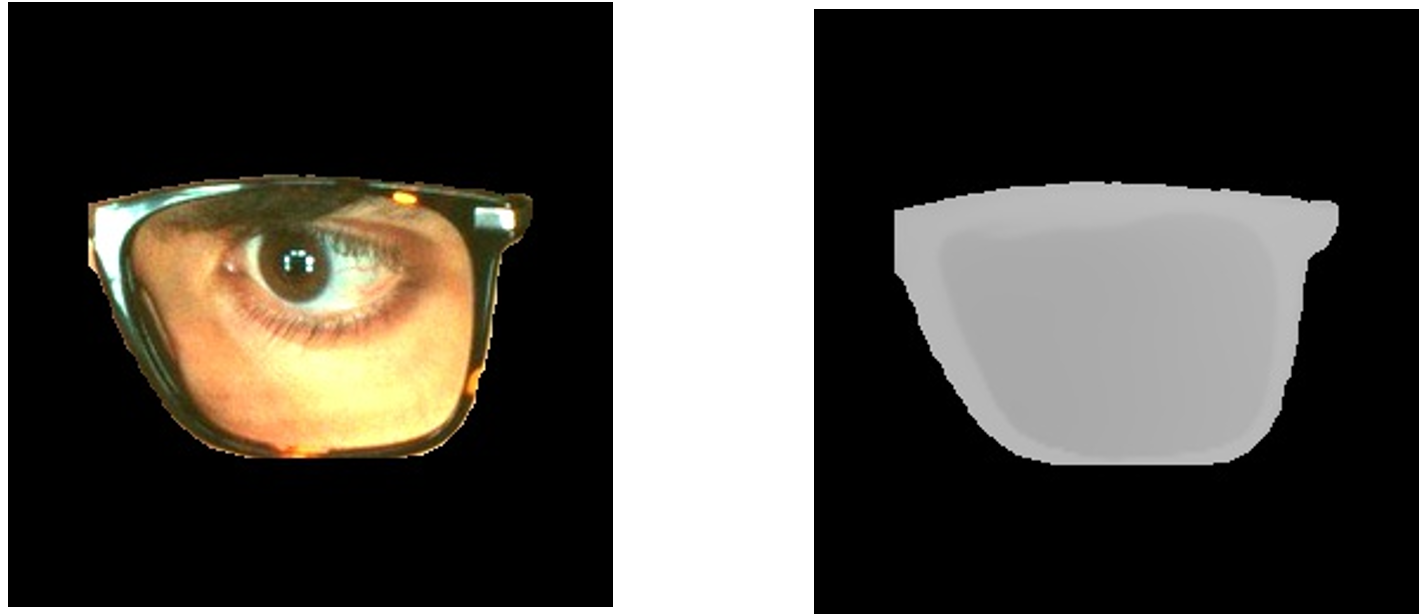}
    \caption{\textbf{Color and depth information to obtain measurements.} The figure shows a particular image acquisition, its corresponding depth information; as well as the same information after applying our segmentation model to infer the region of interest (this represents our input data).}
    \label{fig:fig-input}
\end{figure}

\noindent \textbf{Method.}
In this work, we propose a tracer model that utilizes, as input, a 4D tensor composed of an RGB image concatenated with an additional depth channel, four times, one per image. This design enables the model to leverage both color and geometric information via its depth, thereby facilitating more accurate estimations by exploiting the implicit 3D structure of the object learned from prior data. At this stage, the information obtained from the preceding steps is incorporated, i.e., the segmentation masks and the corresponding depth maps. The segmentation mask is applied uniformly across all four channels, ensuring that only the regions of interest are retained while background and irrelevant areas are removed. This preprocessing step allows the model to focus on the salient portions of the input, thereby reducing background interference and promoting specialization in a more constrained set of features. Consequently, the approach contributes to a more compact model with improved predictive accuracy. Fig.~\ref{fig:fig-input} illustrates an example of the input tensor, comparing the data before and after the application of the segmentation mask on the region of interest. Our system uses this last information as input.

From each sample acquisition, two distinct images are generated, each corresponding to one eye. The rationale behind this strategy is to expand the training dataset while reducing the complexity of the learning problem. Instead of processing the entire image, where the model would be required to infer two separate frame traces simultaneously, the proposed approach feeds the model with a single eye frame per input. Under this assumption, the model is expected to effectively capture the object’s 3D spatial representation.

\begin{figure}[t!]
    \centering
    \includegraphics[width=9cm]{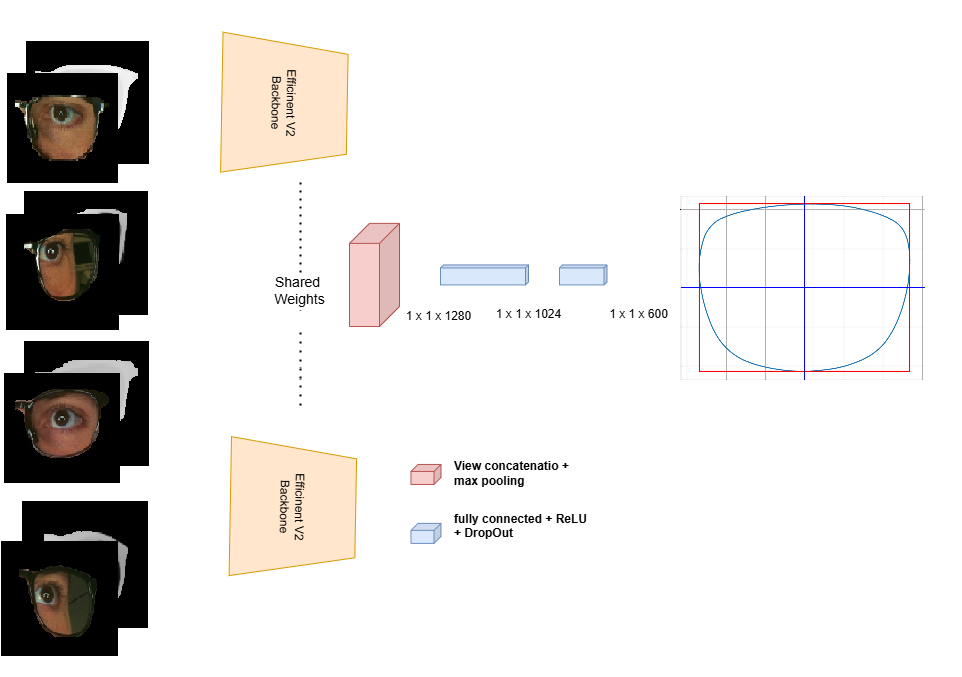}
    \caption{\textbf{Main architecture for eye-frame trace.} Multi-view RGB and depth channels are processed at the same networks.}
    \label{fig:heads-arch}
\end{figure}

MVS models are well suited for this task, as they aggregate information from multiple viewpoints to extract 3D structural features. By leveraging four available image views and a 4D input representation (RGB plus depth) per view, the model is designed to produce accurate 3D measurements. The proposed approach emphasizes two core components within the MVS framework: the feature extraction module and the feature fusion strategy. As a feature extraction module, we take inspiration by EfficientNetV2~\cite{c76} and its variants (S, M, and L), which offer an excellent balance among performance, model size, and computational efficiency. Moreover, the availability of multiple architecture scales allows for a direct comparison across configurations, facilitating the selection of the most suitable variant without structural modification. The overall architecture employs a shared feature extractor network, as illustrated in Fig.~\ref{fig:heads-arch}.

The fusion of features across multiple views constitutes a fundamental component of MVS frameworks. Several aggregation strategies addressing this aspect have been explored in prior work ~\cite{c71}, ~\cite{c102}. Two alternatives are considered: early fusion, and late fusion. In early fusion, feature integration occurs at the convolutional level. Each input view is processed by a shared subnetwork to produce feature maps, which are then concatenated across views. Two aggregation methods can be exploited here: Max-pooling across the view dimension or convolution to learn optimal view combinations. In late fusion, each view is independently processed to obtain feature vectors which are then aggregated using either max-pooling over views or fully connected fusion, which learns weighted combinations of all view features. Late fusion preserves high-level semantic information and introduces fewer additional parameters than early fusion.

\section{Experimental Results}
\label{section_experiments}

Next our experimental evaluation on real data is presented, providing both quantitative and qualitative evaluations in all modules we propose.

\subsection{Segmentation}

We begin by evaluating the segmentation strategy to assess its effectiveness in accurately extracting the regions of interest. As mentioned above, we examine multiple model configurations with varying capacities, denoted as {\em tiny}, {\em small}, {\em base+}, and {\em large}. A comparative summary of the results, including model size, inference speed, and IoU, is presented in Table~\ref{tab:experiment3}.
As shown, the {\em small}, {\em base+}, and {\em large} models yield comparable performance, all significantly outperforming the {\em tiny} configuration. In general, prediction performance improved with larger model sizes, but inference time also increased accordingly. However, due to its high computational cost and memory requirements, the {\em large} model is excluded from further consideration. Balancing IoU against model size and inference efficiency, both {\em small} and {\em base+} variants are selected for fine-tuning on our dataset.

\begin{figure*}[t!]
    \centering
    \resizebox{16.8cm}{!} {
\begin{tabular}{@{}ccccc@{}}
    \includegraphics[clip, angle=0,width=1.0\linewidth]{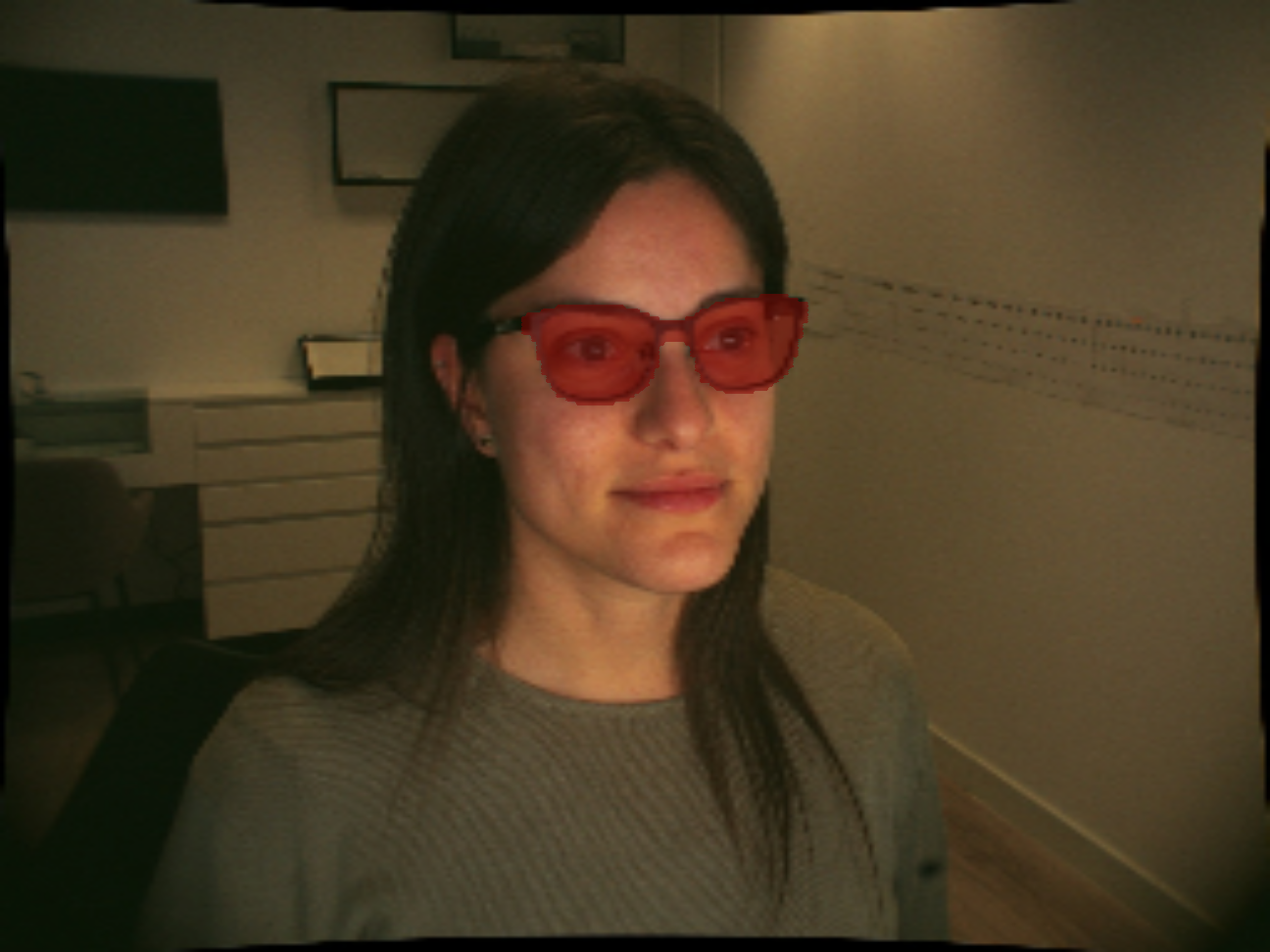}&
     \includegraphics[clip, angle=0,width=1.0\linewidth]{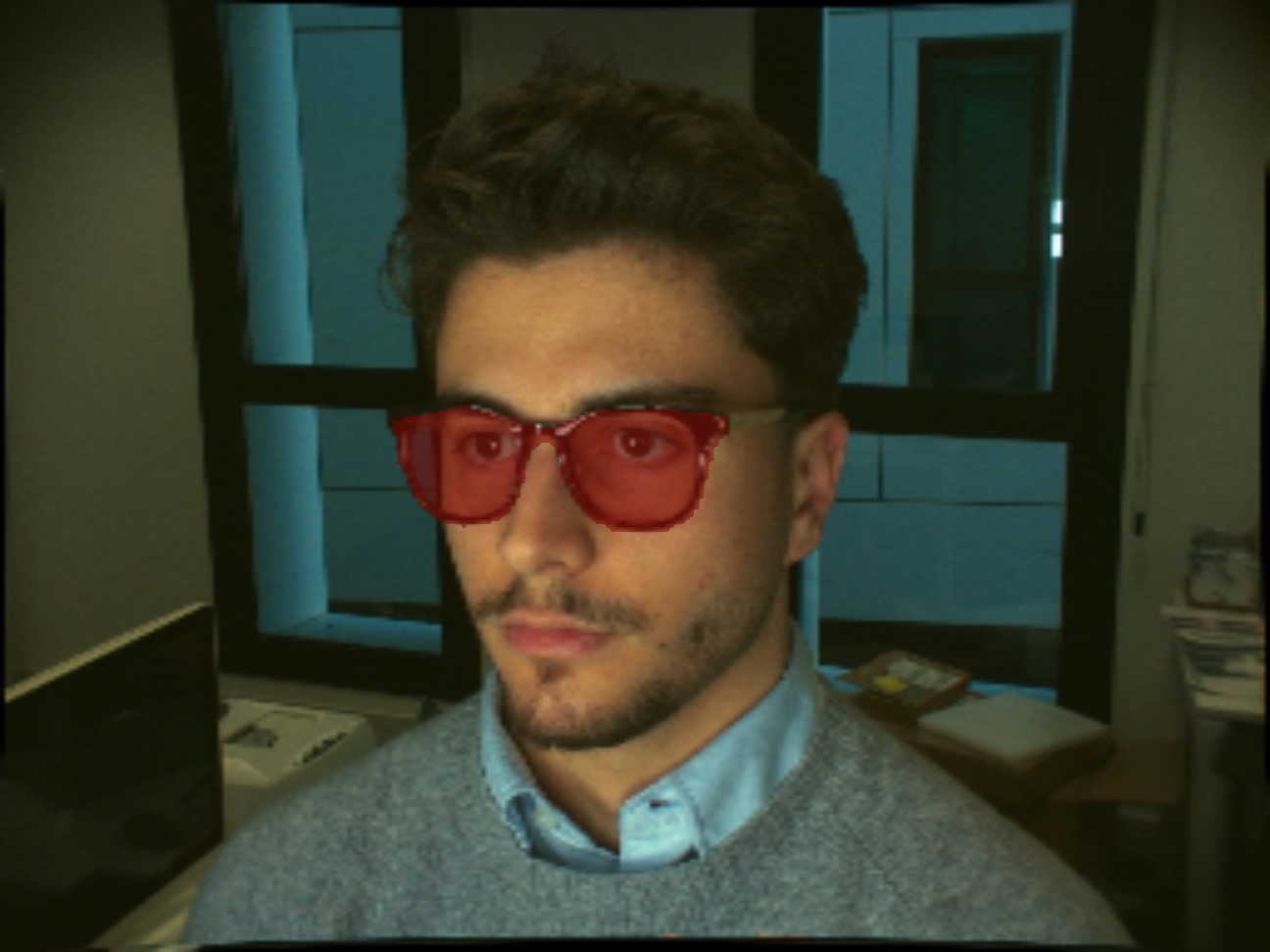}&
     \includegraphics[clip, angle=0,width=1.0\linewidth]{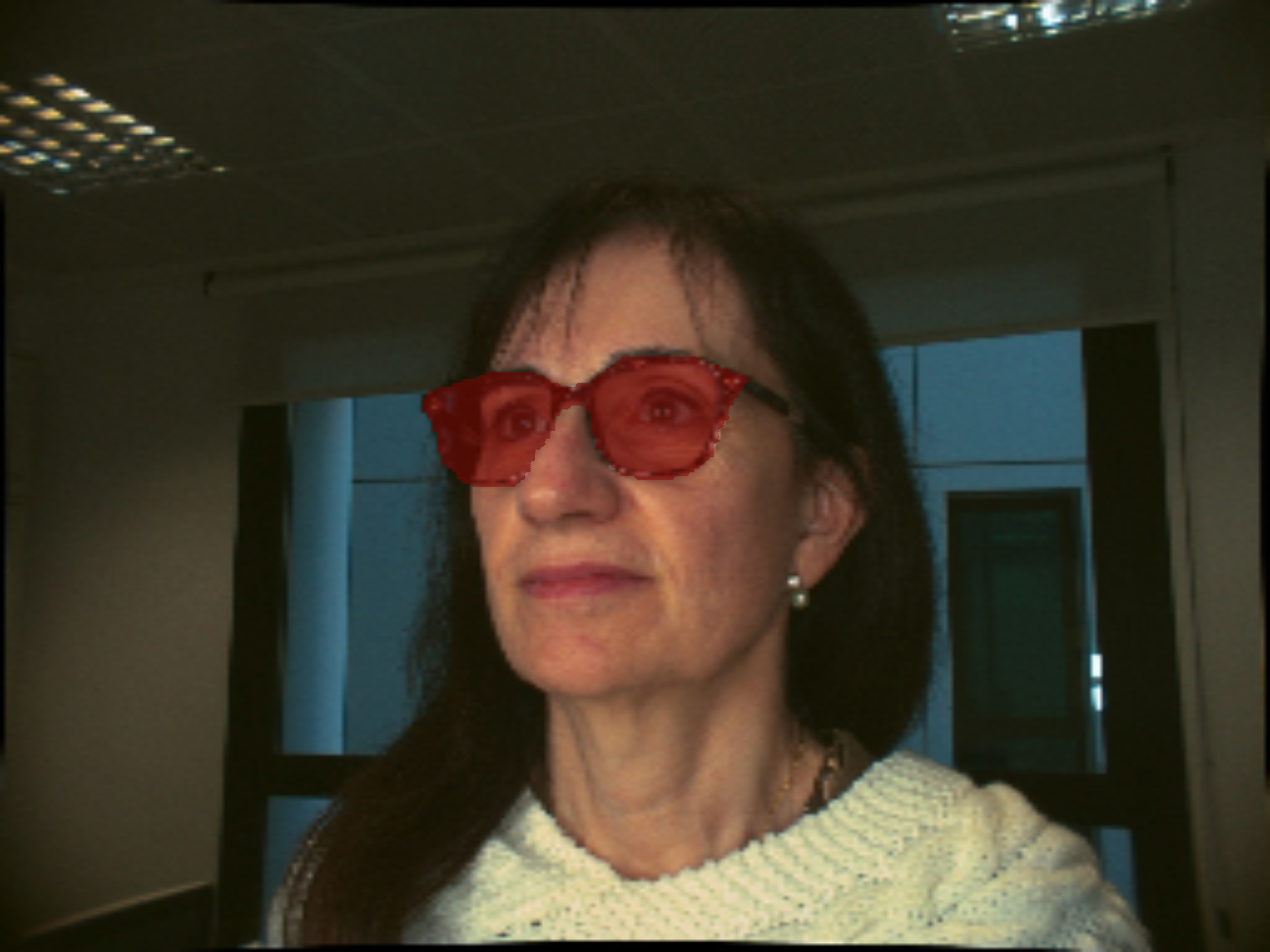}&
     \includegraphics[clip, angle=0,width=1.0\linewidth]{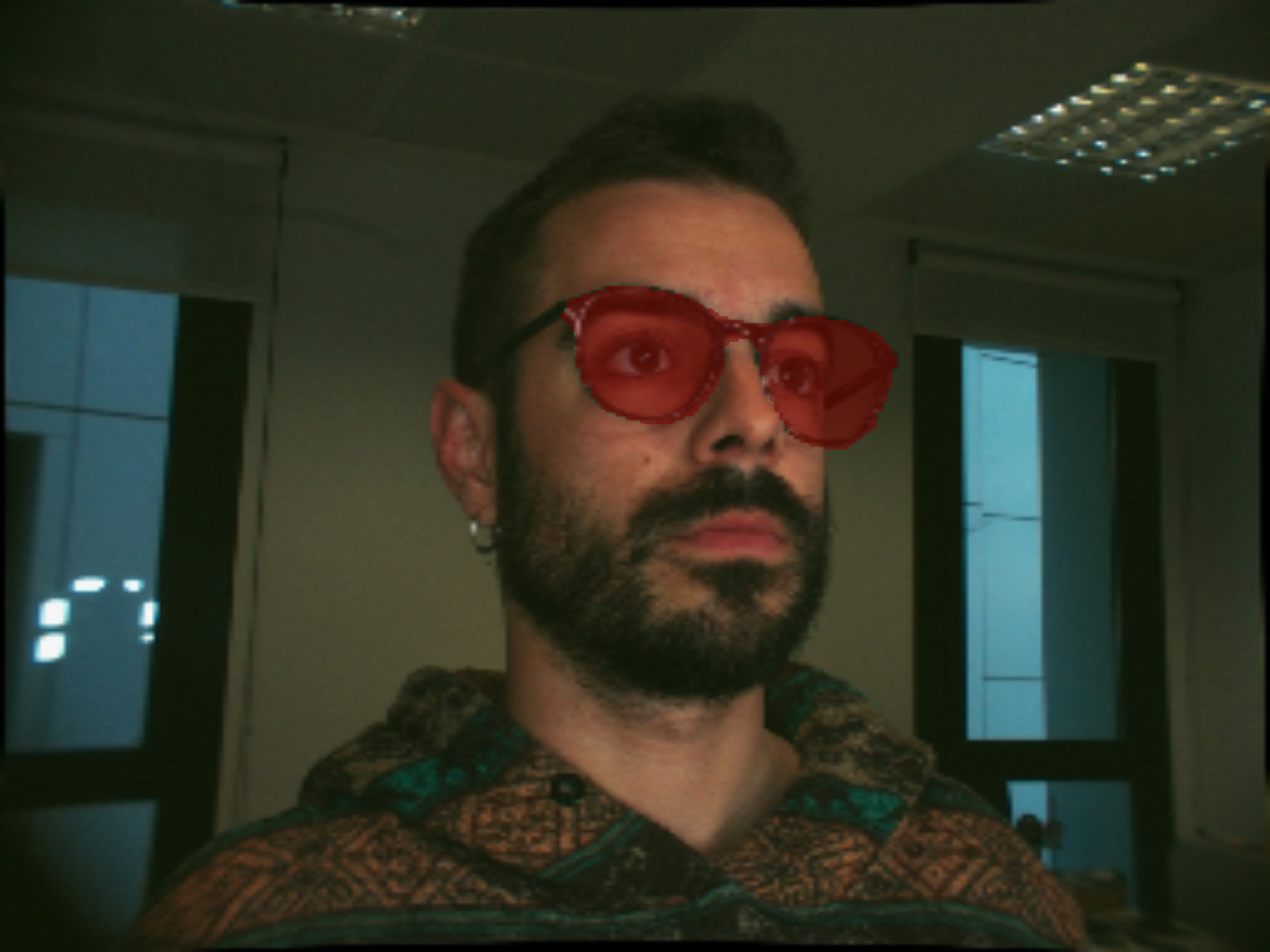}&
     \includegraphics[clip, angle=0,width=1.80\linewidth]{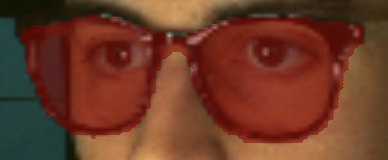}\\
    \includegraphics[clip, angle=0,width=1.0\linewidth]{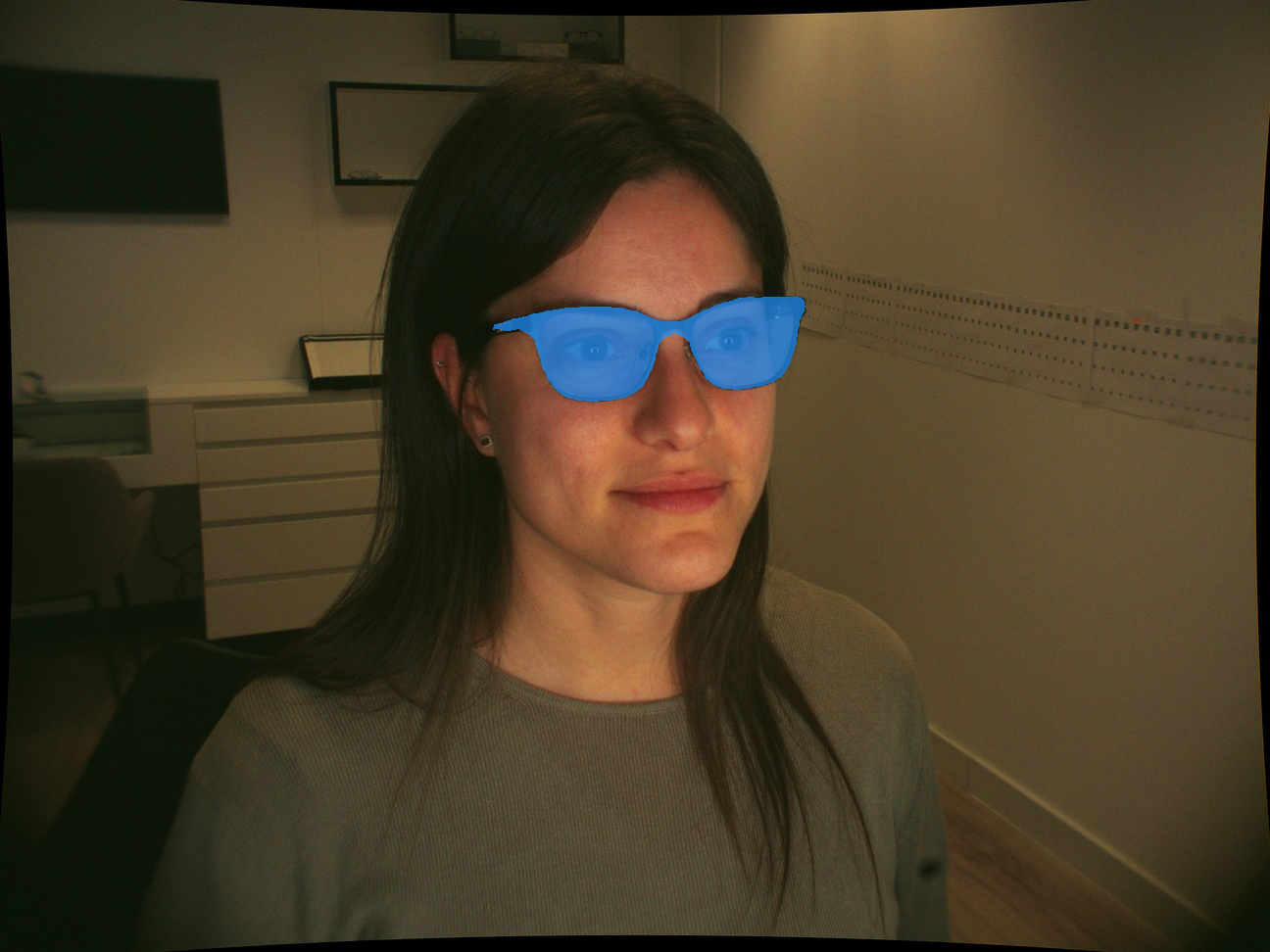}&
    \includegraphics[clip, angle=0,width=1.0\linewidth]{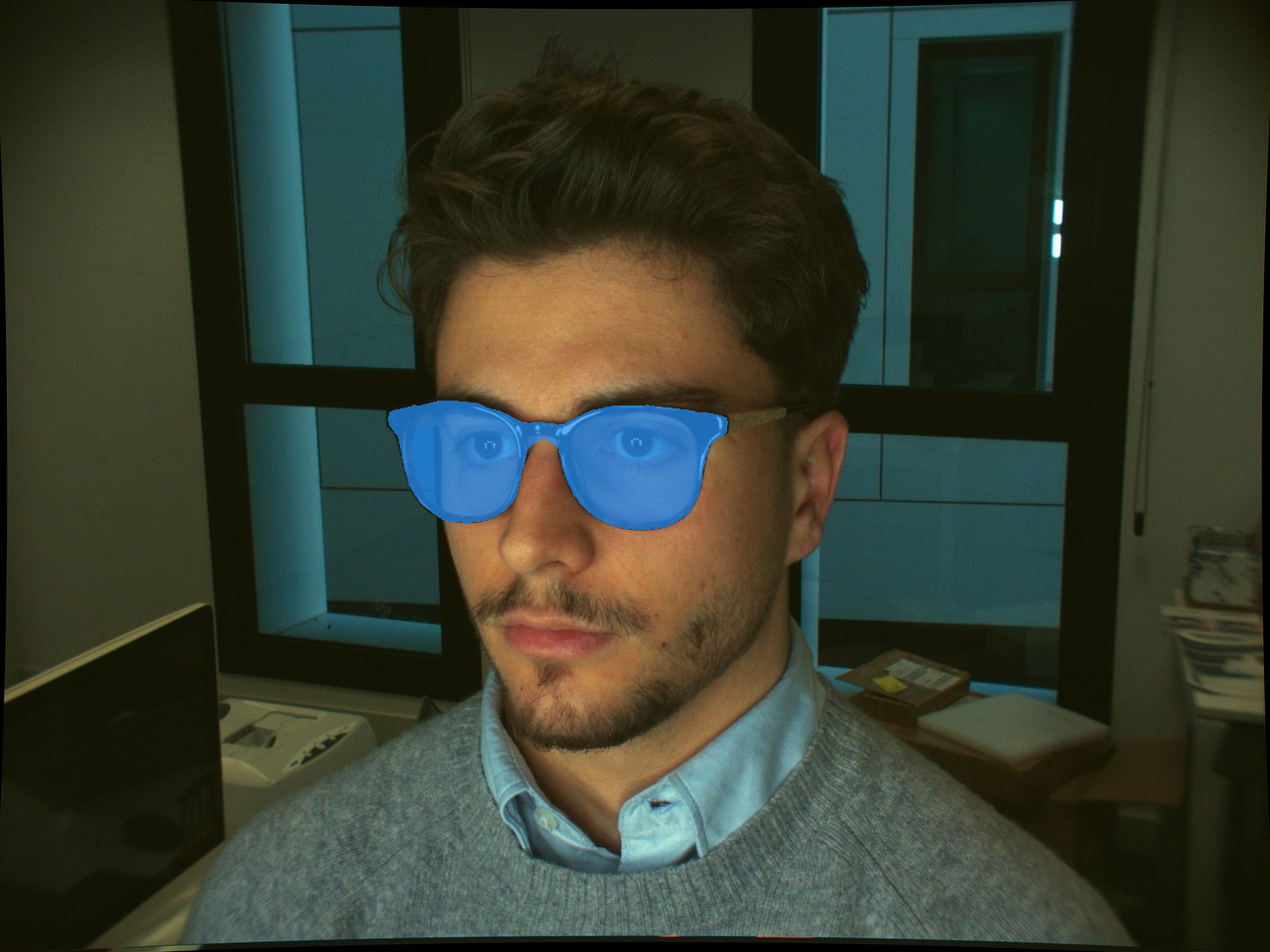}&
    \includegraphics[clip, angle=0,width=1.0\linewidth]{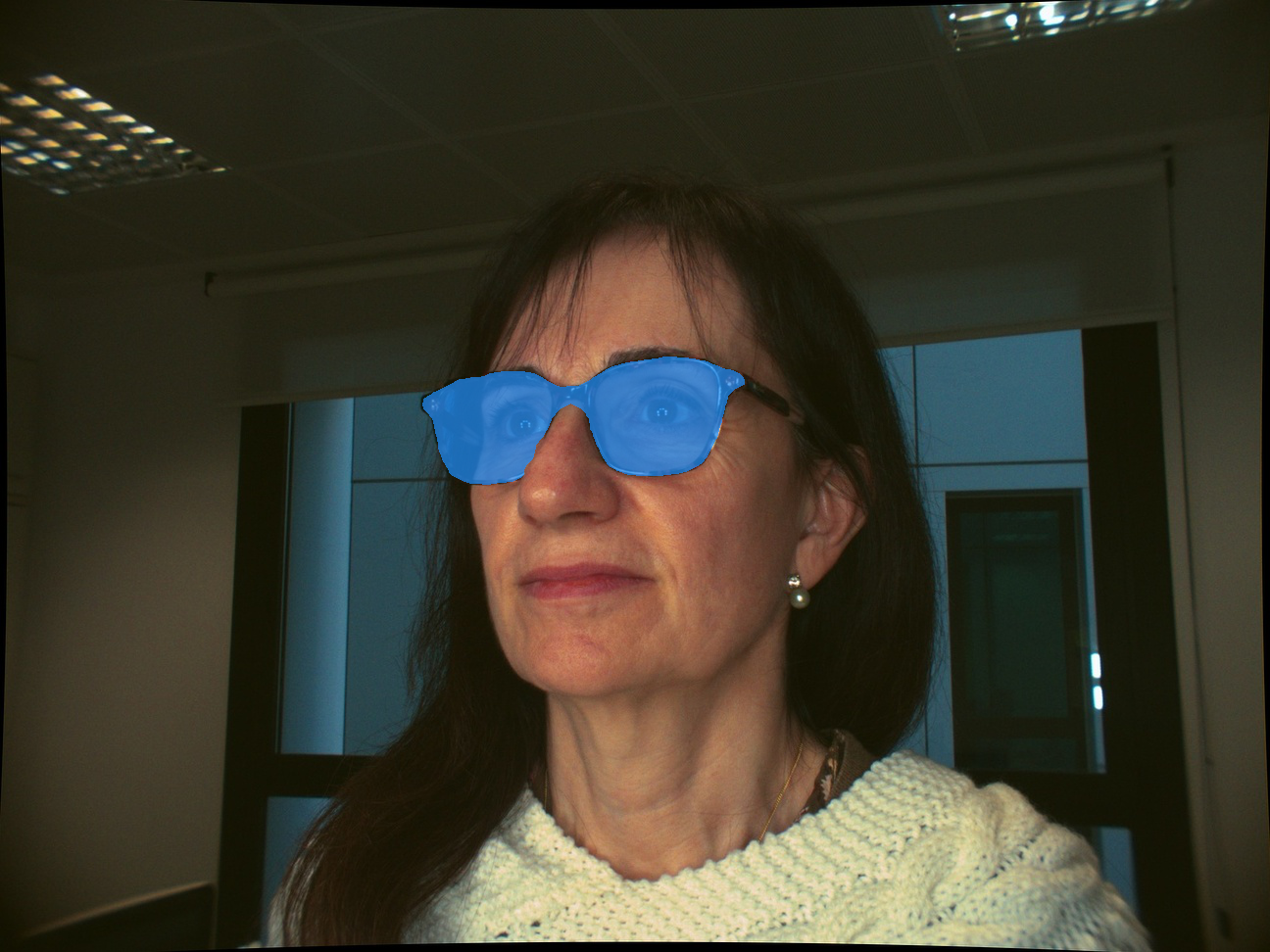}&
    \includegraphics[clip, angle=0,width=1.0\linewidth]{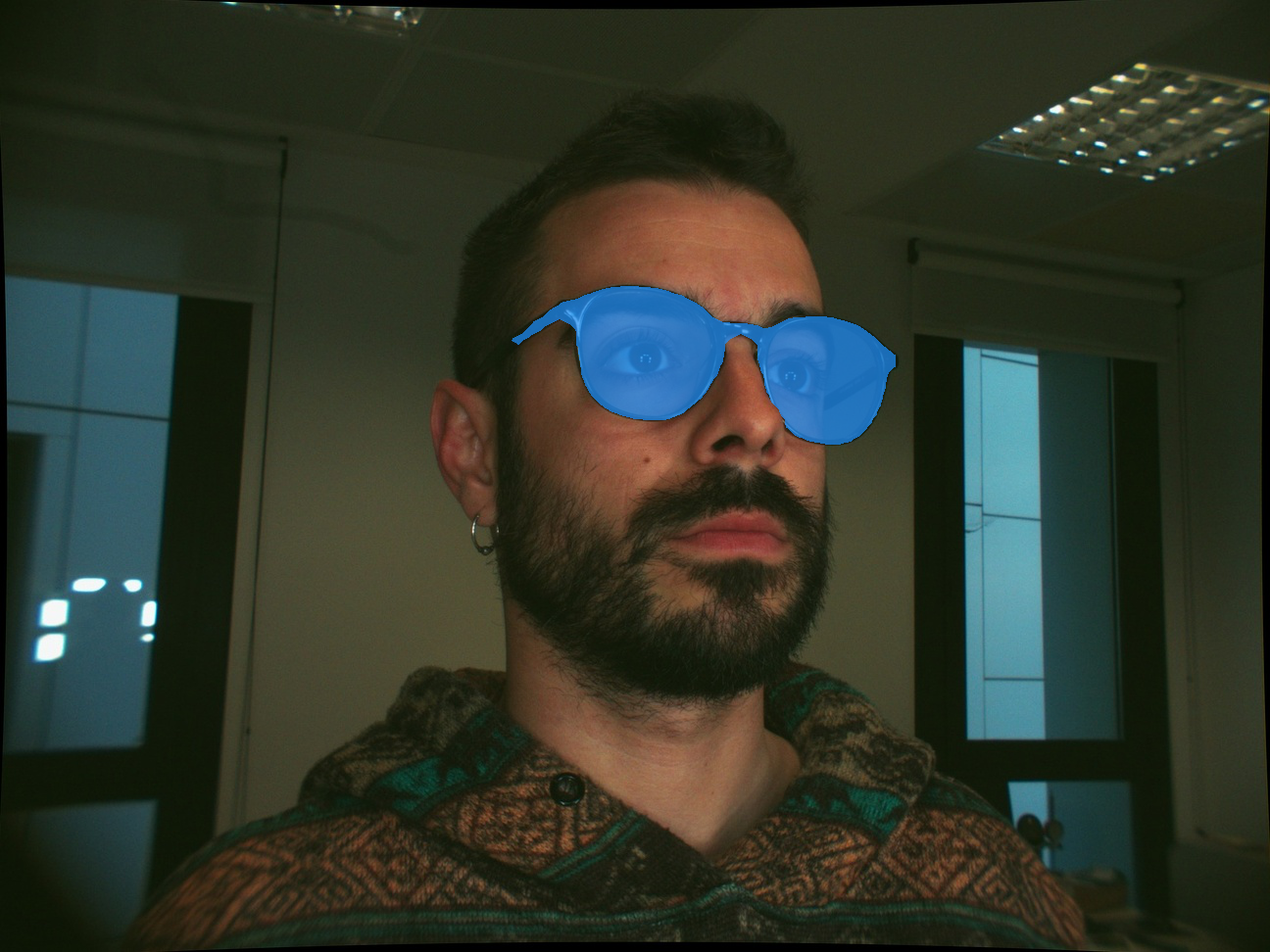}&
     \includegraphics[clip, angle=0,width=1.80\linewidth]{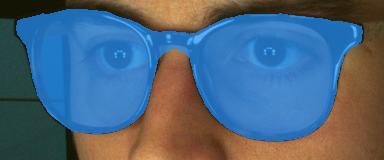}
    \end{tabular}}
    \caption{\textbf{Qualitative evaluation of our segmentation strategy with the {\em base+} model for different users.} The glass segmentation applying DeepLabv3+~\cite{c20} and ours is displayed in red and blue, respectively. In last column, a zooming plot is shown. Best viewed in color.} 
    \label{fig:fig11}
\end{figure*}

To this end, the segmentation model was fine tuned on a total of 1,002 images with 1024$\times$1024 pixels: 752 for training, 100 for validation and 150 for testing. A batch size of 4 and an Adam optimizer with a learning rate of 0.0001 are used. The loss used is a mix of Cross entropy loss and IoU loss. After finetuning, the updated results are summarized in Table~\ref{tab:segmentation-results}. As shown, the {\em base+} configuration achieves the highest performance, reaching an IoU of 0.958, an outstanding outcome for a few-shot segmentation scenario. Representative examples of this configuration are illustrated in Fig.~\ref{fig:fig11}-bottom, highlighting the high segmentation quality attained by the proposed method. In fact, this solution is more accurate with respect to that provided by DeepLabv3+~\cite{c20} (see last column in the figure). Once the segmentation is inferred, a final mask per eye can be easily extracted. It is worth noting that there is some overlap with the user's nose in some cases (see fourth column in the figure), although this rarely happens and therefore does not drastically affect our solution. These masks are finally adapted to the resolution of our system, that provided in the inVision system.

\begin{table}[t!]
\centering
\caption{\textbf{Segmentation evaluation for several configurations.} The table reports model size (in Mb), inference speed (in Hz), and IoU, for different configurations.}
\begin{tabular}{|l|c|c|c|c|}
\hline
\text{Metric} & \text{Tiny} & \text{Small} & \text{Base+} & \text{Large} \\ \hline
Size [Mb] $\downarrow$     &  \textbf{38.9 }          & 46.0   &  80.8              & 224.4  \\ \hline
Speed [Hz]  $\uparrow$   & \textbf{47.2 }           & 43.3   &  34.8            & 24.2  \\ \hline
IoU   $\uparrow$        & 0.902            & 0.922   &  0.951            & \textbf{0.961}  \\ \hline
\end{tabular}
\label{tab:experiment3}
\end{table}

\begin{table}[t!]
\centering
\caption{\textbf{Final Segmentation results after finetuning.} Three configurations are finetuned on the custom dataset. The table reports the train and test losses as well as the corresponding IoU.}
\begin{tabular}{|l|c|c|c|}
\hline
\text{Metric} & \text{DeepLabV3~\cite{c20}} & \text{Small} & \text{Base+}  \\ \hline
Train Loss (Dice loss)  $\downarrow$   & 0.367  & 0.362               & \textbf{0.348}  \\ \hline
Test Loss (Dice loss)  $\downarrow$   & 0.367   & 0.363                 & \textbf{0.348}  \\ \hline
IoU         $\uparrow$      & 0.931          & 0.938             & \textbf{0.958}  \\ \hline
\end{tabular}
\label{tab:segmentation-results}
\end{table}

\subsection{Depth estimation}

We now evaluate the depth estimation module that was introduced above. Because 3D ground truth is not available, we just report a qualitative evaluation.

As was introduced, three variants are considered as a function of the encoder size: ViT-S, ViT-B, and ViT-L. For this experiment, we use the next configuration: 
\begin{itemize}
    \item The model is tested with images of 518$\times$518. They are also normalized with standard normalization.
    \item The model weights were loaded from a pretrained version provided by~\cite{c72}, jointly trained on 1.5M and 62M labeled and unlabeled images, respectively.
    \item The output depth values are normalized between 0 and 255.
\end{itemize}

\begin{figure*}[t!]
\centering
\resizebox{16.8cm}{!} {
\begin{tabular}{@{}cccc@{}}
  \includegraphics[clip, angle=0,width=1.0\linewidth]{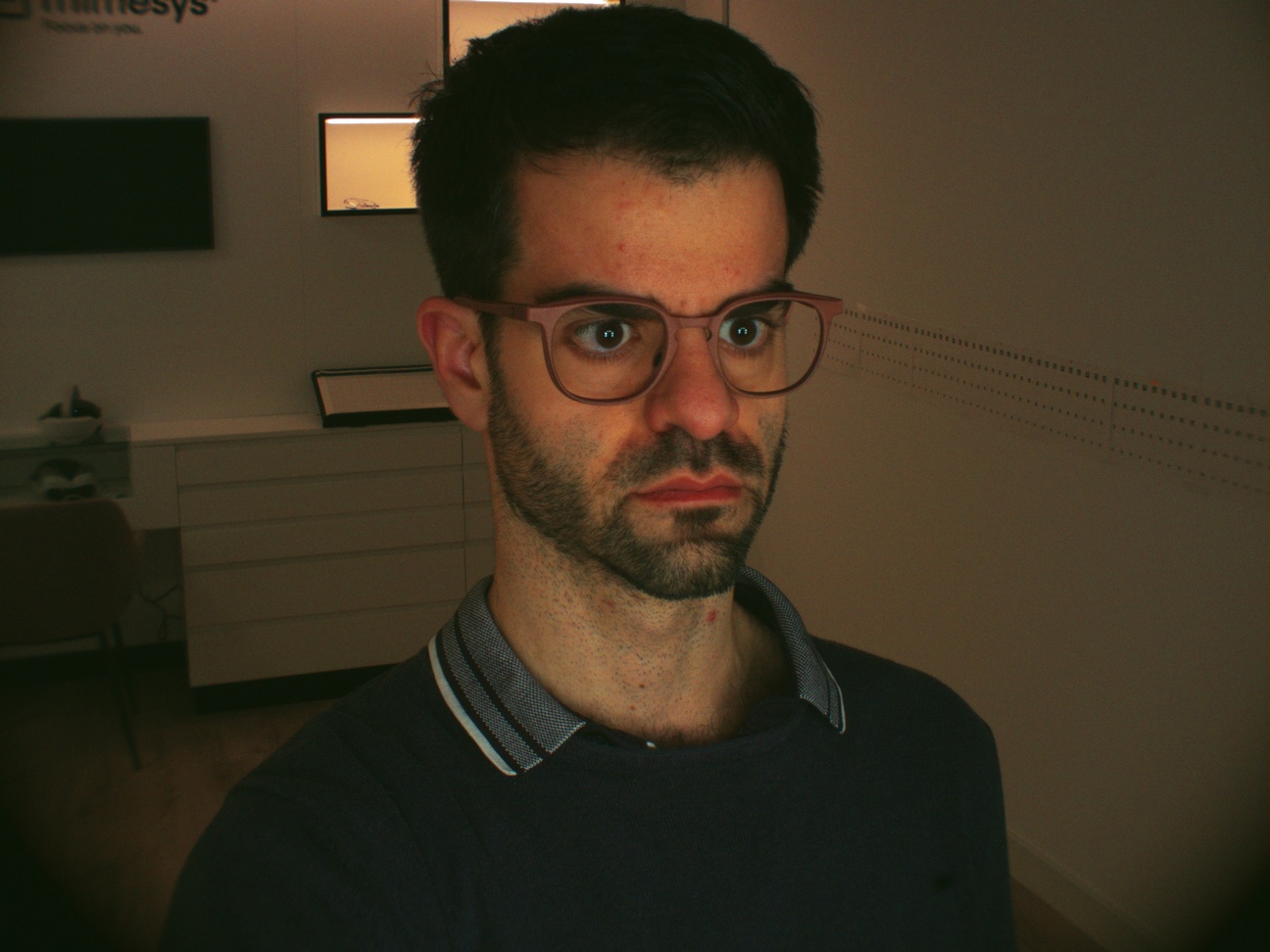}&
    \includegraphics[clip, angle=0,width=1.0\linewidth]{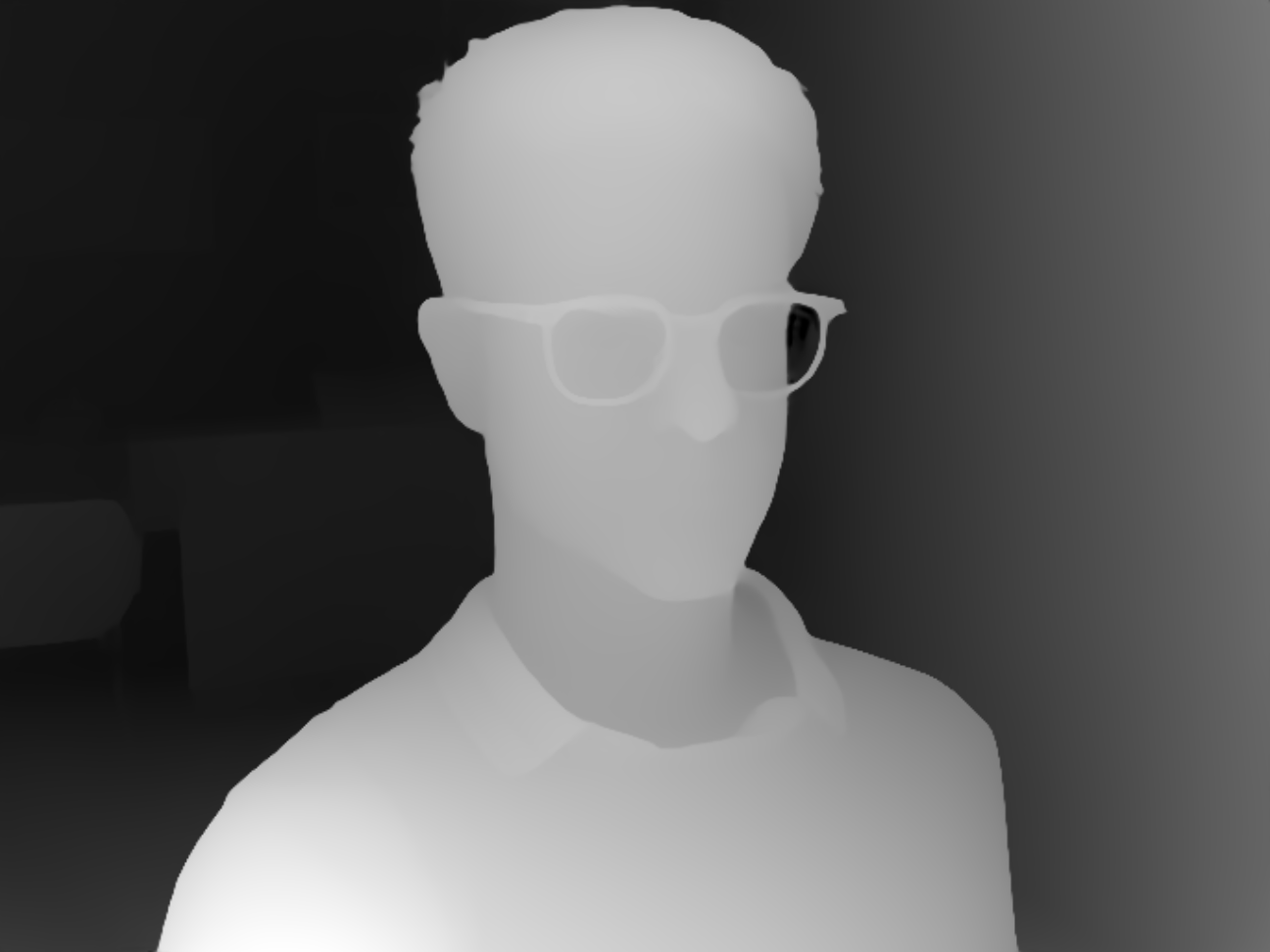}&
      \includegraphics[clip, angle=0,width=1.0\linewidth]{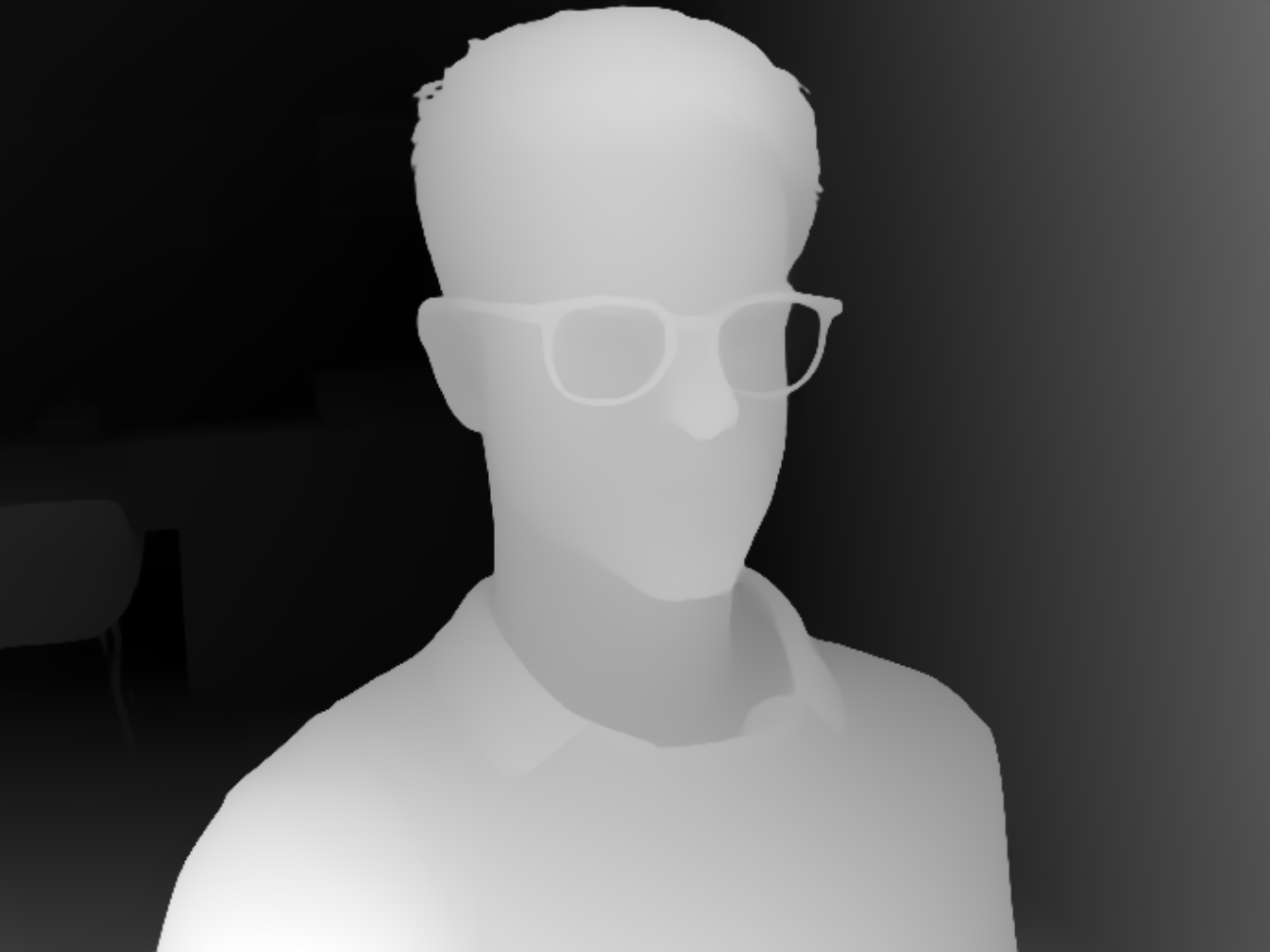}&
        \includegraphics[clip, angle=0,width=1.0\linewidth]{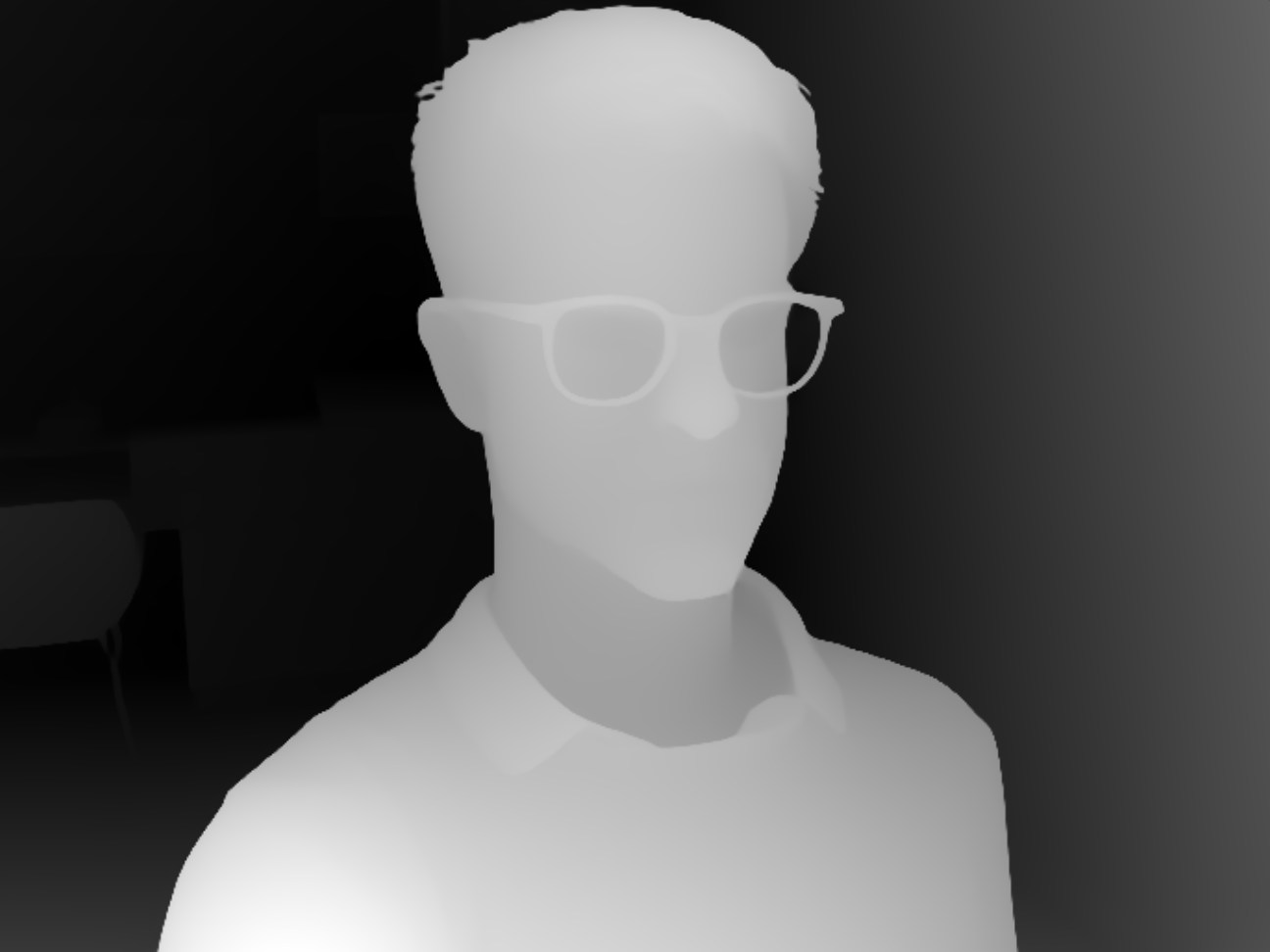}
  \end{tabular}}
    \caption{\textbf{Qualitative results on depth estimation.} From left to right: Test input image, depth prediction using ViT-Small, ViT-Base, and ViT-Large, respectively.}
    \label{fig:deepanything-depthvalues}
\end{figure*}

A qualitative evaluation is displayed in Fig.~\ref{fig:deepanything-depthvalues}, where the foreground details appear consistent across all three encoders. However, significant visual differences appear in the background elements, particularly visible in the chair leg. This is especially relevant between the small model and the larger one. As a consequence, we select the ViT-Base model as the optimal solution, considering a trade off between accuracy and cost (in terms of computational resources) and assuming the foreground is the most important part in the image. As can be seen, the relative depth map is smooth, a key factor in our problem, as we will exploit this information to constrain our neural model and learn to infer accurate trace measurements, especially in the eyeframe region.

\subsection{Trace measurement}

Finally, we present our experimental results on trace measurement. To enable quantitative evaluation, we constructed a new dataset consisting of 500 acquisitions. Among these, 400 samples are allocated for training, 50 for validation, and 50 for testing. For every sample, 4 images are considered, containing everyone of them a 4D tensor that comprises the RGB and depth information, along with the corresponding trace measurement. For our experiments, EfficientNetV2~\cite{c76} is used as a backbone model that is then trained on the ImageNet-1K~\cite{c63} dataset and tested with different sizes: S, M and L with 21M, 54M and 118M of parameters, respectively. A batch size of 8 images at a time and an Adam optimizer with a learning rate of 0.0001 are employed.

\begin{table*}[t!]
\centering
\small  
\setlength{\tabcolsep}{4pt}  
\caption{\textbf{Quantitative evaluation on frame tracer for several modalities and model size.} The table reports the minimum, maximum and median error values of frame radius measurement, all of them in millimeters. The one-head approach is considered.}
\begin{tabular}{|l|c|c|c|}
\hline
\diagbox[width=6em]{Size}{Modality} & \text{RGB (no seg.)}  & \text{Grey + Depth} & \text{RGB + Depth}  \\ \hline
S     & 0.0003 2.8451 0.8910  & \textbf{0.000010 1.8466 0.4238}  & 0.00039 2.3461 0.4881  \\ \hline
M     & 0.0003 3.2052 0.9102  & 0.000014 3.0918 0.4958  & 0.00087 2.1069 0.4818  \\ \hline
L     & 0.0005 3.8454 0.9686 & 0.000052 3.5024 0.5834  & 0.00012 3.2422 0.4840   \\ \hline
\end{tabular}
\label{tab:experiment-mvdcnn}
\end{table*}

First, we evaluate several input modalities (RGB without segmentation, grayscale combined with depth information, and RGB combined with depth) for each proposed model size. In this case, the latest max-pooling fusion strategy is used. The corresponding results are presented in Table~\ref{tab:experiment-mvdcnn}. The first column reports the performance of a vanilla network, defined as an EfficientNetV2~\cite{c76} backbone followed by a fully connected regressor, without incorporating either segmentation or depth cues. In this setting, the best configuration corresponds to the ``S'' model, which attains average and maximum errors of 0.8910 mm and 2.8451 mm, respectively. In contrast, the overall best performance is achieved by the smallest backbone trained using the grayscale+depth modality. This configuration yields a mean error of 0.4238 mm and a maximum one of 1.8466 mm across all frames, representing more than a 50\% reduction in mean error compared to the vanilla model. In the proposed design, the early fusion of RGB and depth channels disrupts the initial layers and their pretrained representations, whereas the single-channel depth and grayscale modalities exhibit greater compatibility due to their structurally similar one-channel format. Furthermore, we can see that as the model size increases, the trace performance worsens. Consequently, we will select the smallest model along with the grayscale+depth modality.

\begin{table}[t!]
\centering
\caption{\textbf{Fusion Strategy Evaluation for trace measurements.} The table reports the frame tracer results for different fusion strategies. Minimum, maximum and median error values of frame radius measurement are included, all of them in millimeters. The one-head approach and model ``S'' are considered.}
\begin{tabular}{|l|c|}
\hline
\diagbox[width=5.5em]{Strategy}{Model} & \text{Gray + Depth }  \\ \hline
Early Fusion max-pooling  & 0.00002 2.0108 0.4433  \\ \hline
Early Fusion 1$\times$1 conv    & 0.00010 2.1253 0.4437  \\ \hline
Late Fusion fc          & 0.00002 1.8738 0.4517   \\ \hline
Late Fusion max-pooling & \textbf{0.00001 1.8466 0.4238} \\ \hline
\end{tabular}
\label{tab:experiment-mvdcnn2heads}
\end{table}

We now evaluate several fusion strategies for multi-view features for model ``S'' and the grayscale+depth modality, the best one in our previous analysis. In particular, we experiment the fusion strategy: early fusion at the penultimate layer versus late fusion; and the fusion methodology (max-pooling fusion versus learnable fusion through 1$\times$1 convolutions and fully connected networks). The results are reported in Table~\ref{tab:experiment-mvdcnn2heads}. As can be seen, the best fusion strategy is later fusion max-pooling, enhancing the performance of our multi-view image tracer system for the EfficientNetV2~\cite{c76} backbone approach with model ``S''. This combination obtains a minimum, maximum and mean error of 0.00001 mm, 1.8466 mm and 0.4238 mm, respectively, a very competitive measurement for frame trace. In fact, this solution increases performance by 53\% compared to the vanilla solution. In general, our method produces 88\% of measures under 1 mm, i.e., is effective to produce more accurate and robust solutions. Unfortunately, no camera-based solutions for eyeframe measurement were found in the literature for comparison. However, we can provide an indirect comparison with some works. Solutions in metrology, such as~\cite{c12}, used three different neural networks for multi-frequency phase-shifting profilometry, obtaining an error close to 2 mm. In~\cite{c101} was proposed a method to improve the profilometry-based measurements in shadow captures, obtaining a mean error of 1.0279 mm. Despite providing good numbers, our alternative seems more accurate and easier to apply.

\begin{figure*}[t!]
    \centering
    \resizebox{16.8cm}{!} {
        \begin{tabular}{@{}cccc@{}}
          \includegraphics[clip, angle=0,width=1.0\linewidth]{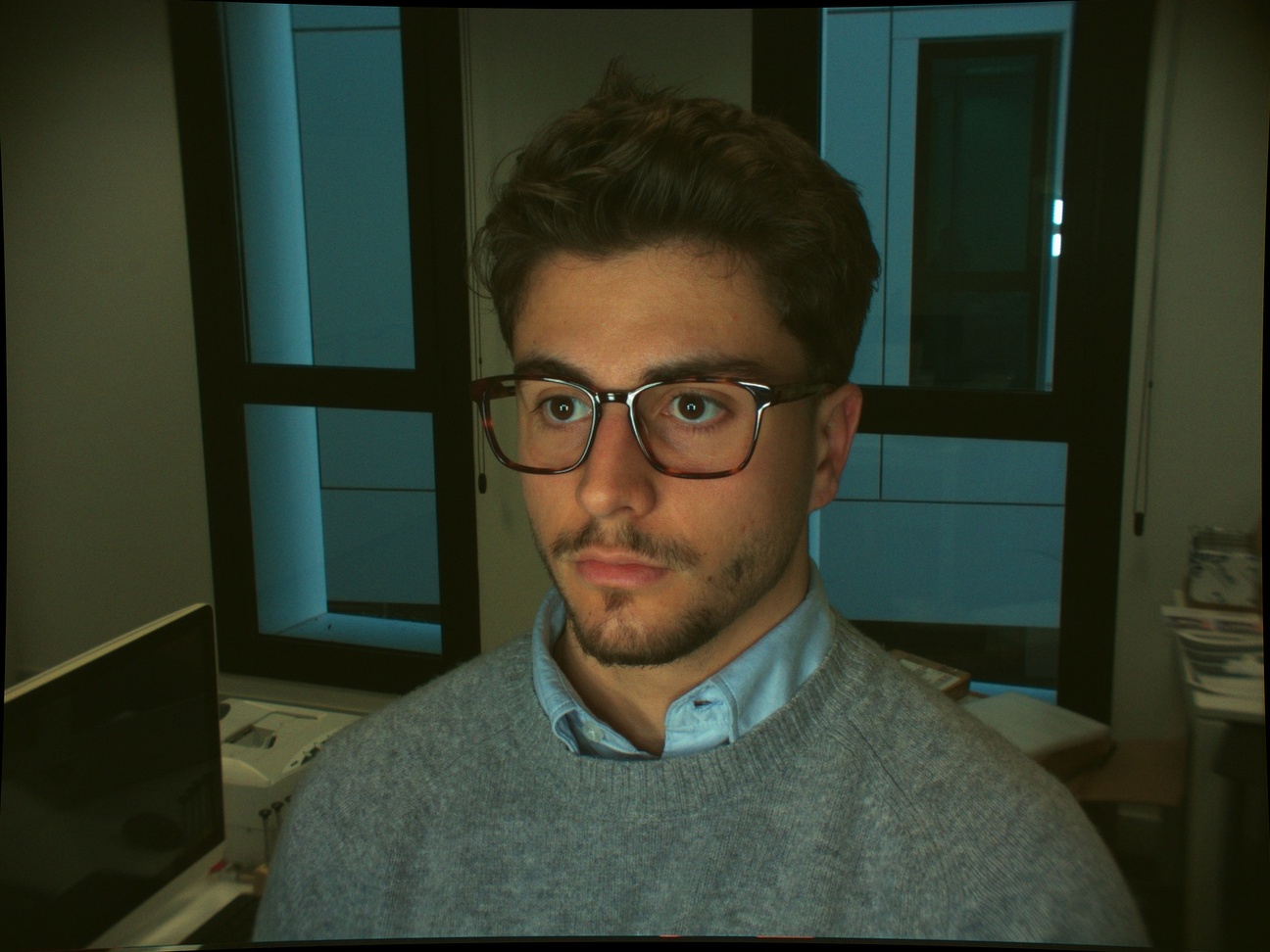}&
            \includegraphics[clip, angle=0,width=1.0\linewidth]{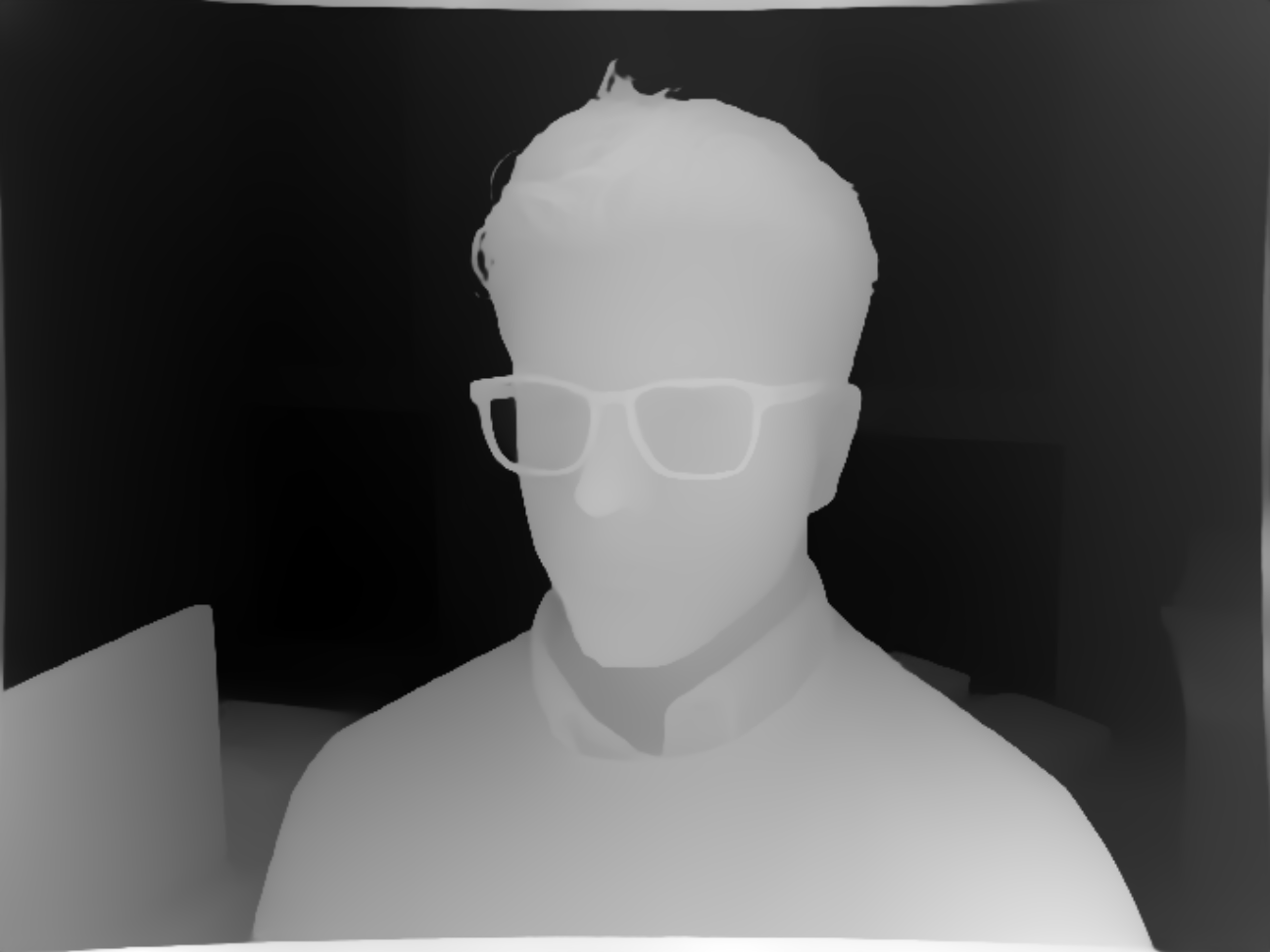}&
              \includegraphics[clip, angle=0,width=1.0\linewidth]{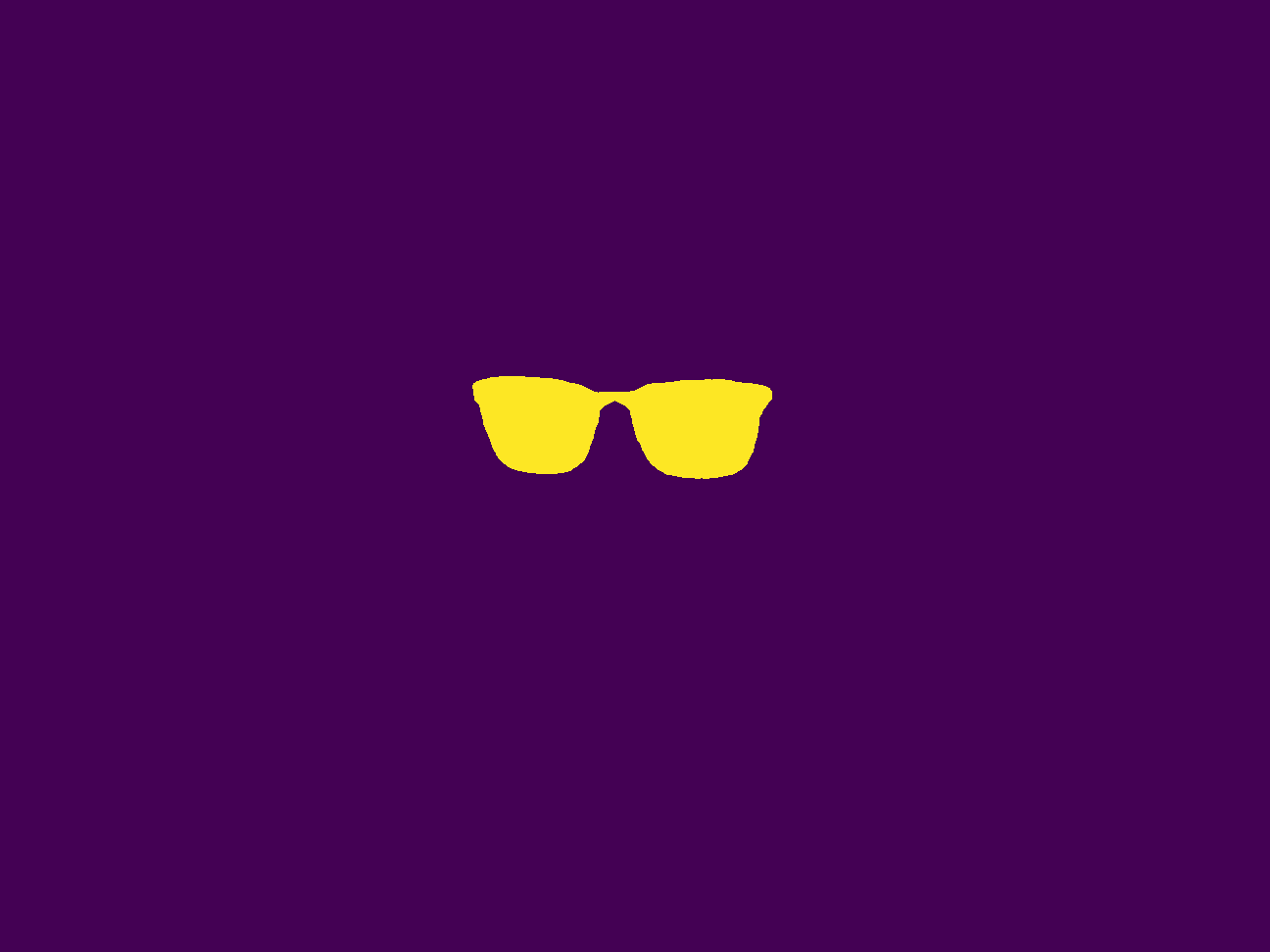}&
                \includegraphics[clip, angle=0,width=1.0\linewidth]{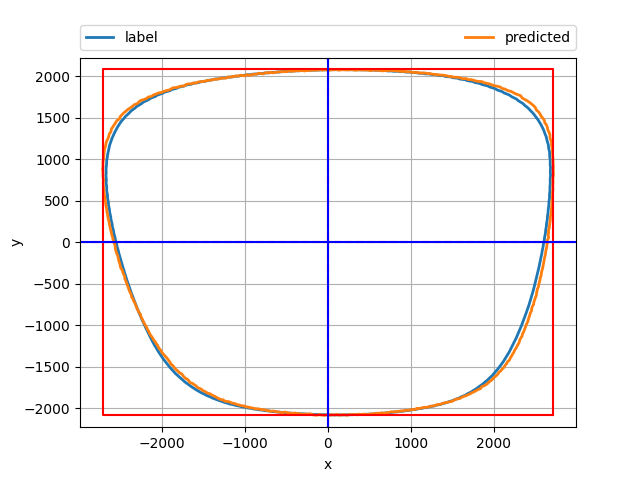} \\
          \includegraphics[clip, angle=0,width=1.0\linewidth]{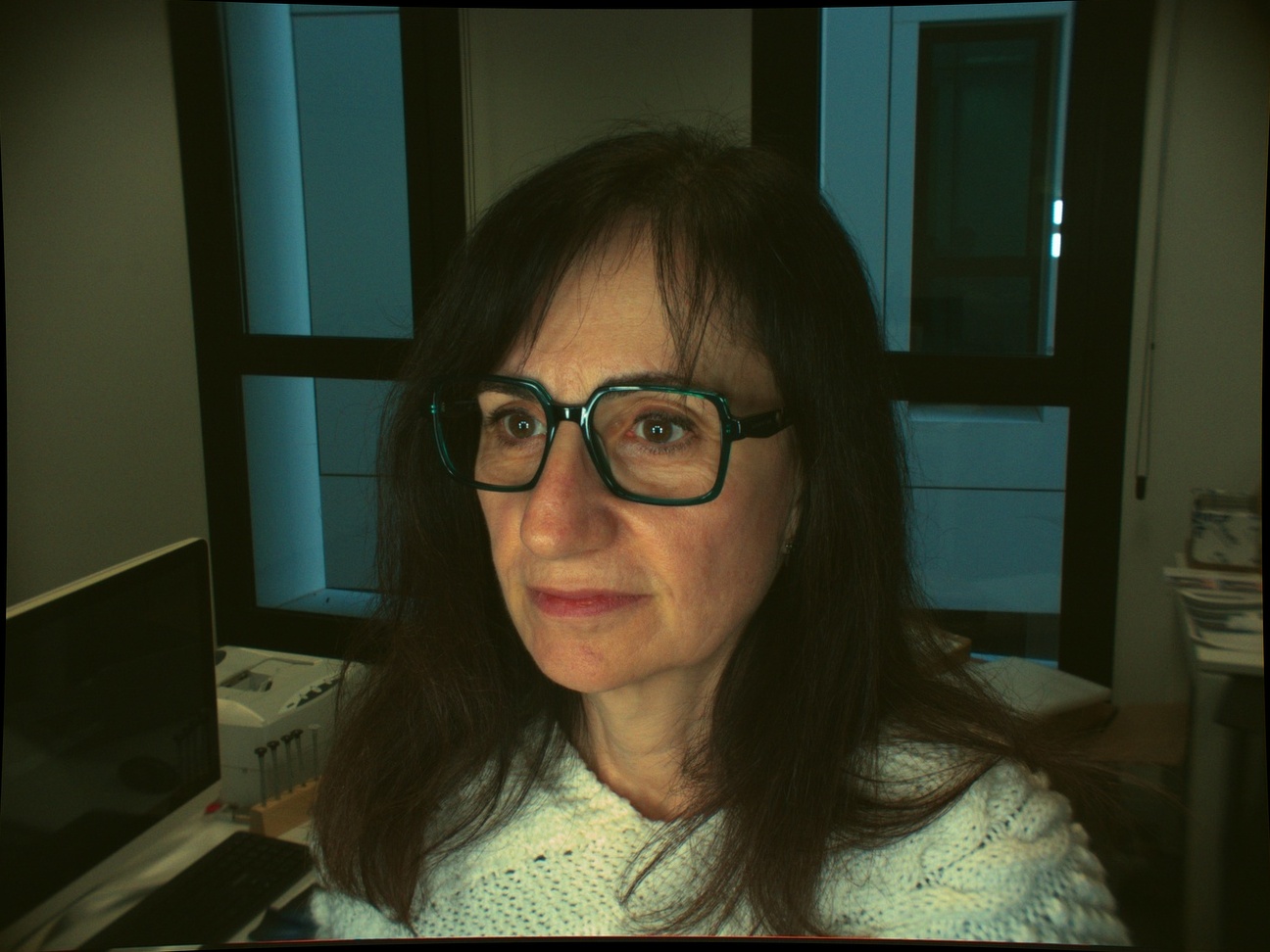}&
            \includegraphics[clip, angle=0,width=1.0\linewidth]{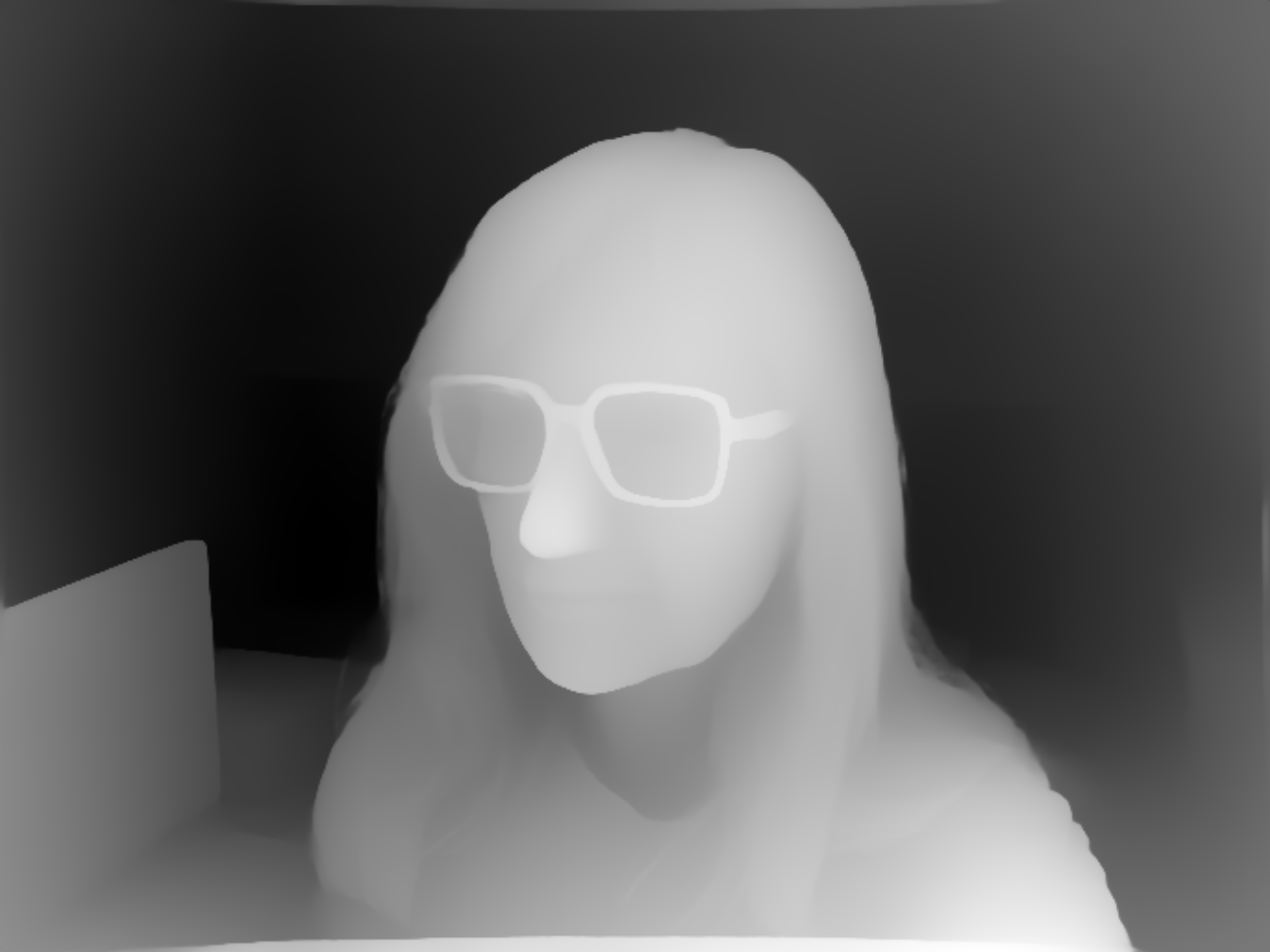}&
              \includegraphics[clip, angle=0,width=1.0\linewidth]{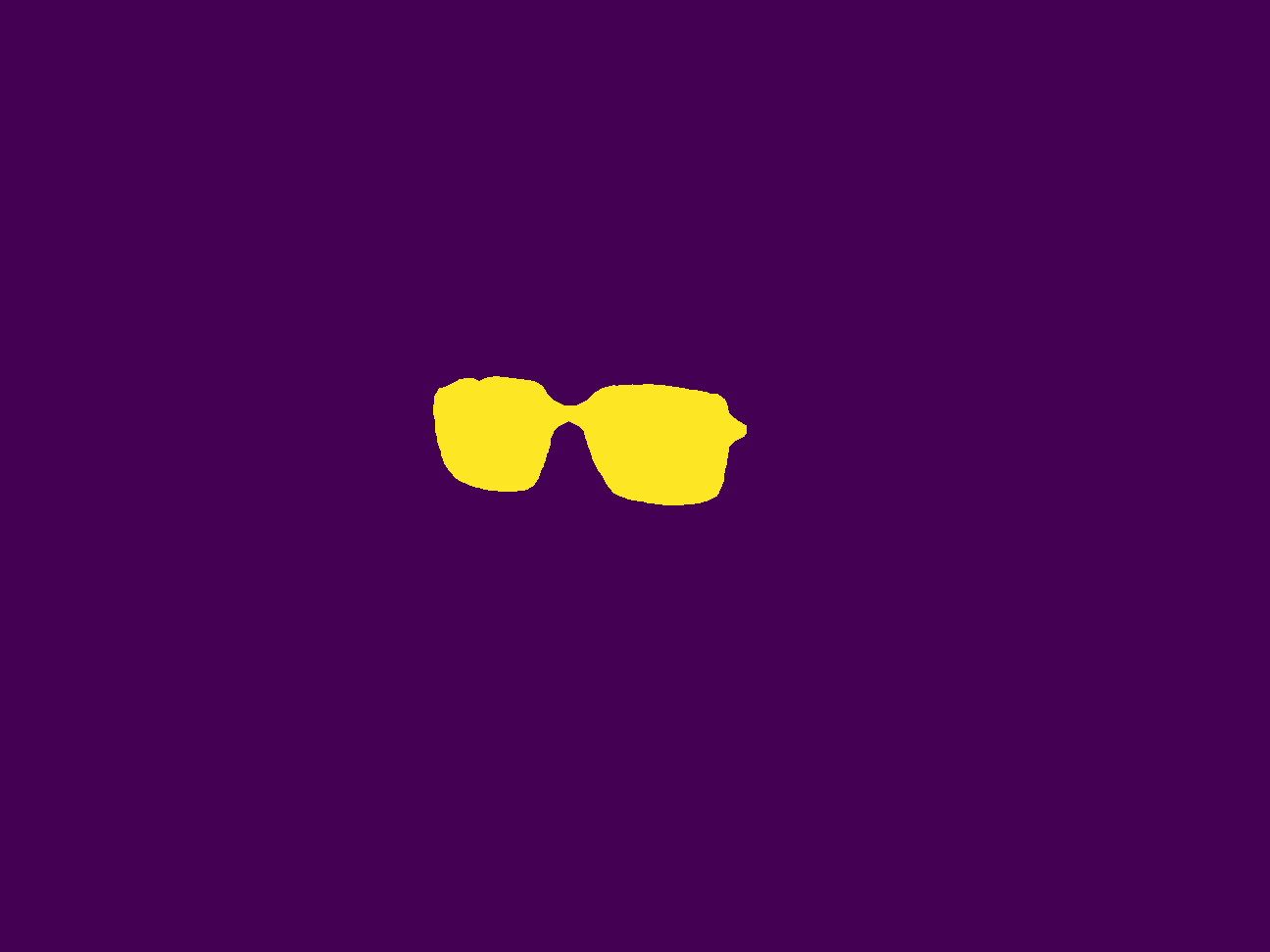}&
                \includegraphics[clip, angle=0,width=1.0\linewidth]{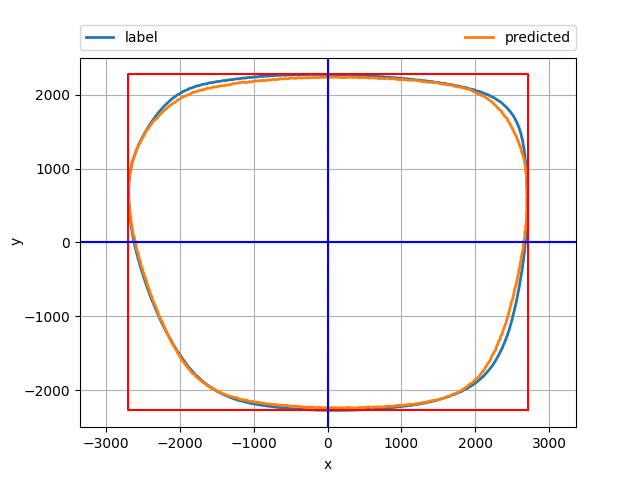}\\
          \includegraphics[clip, angle=0,width=1.0\linewidth]{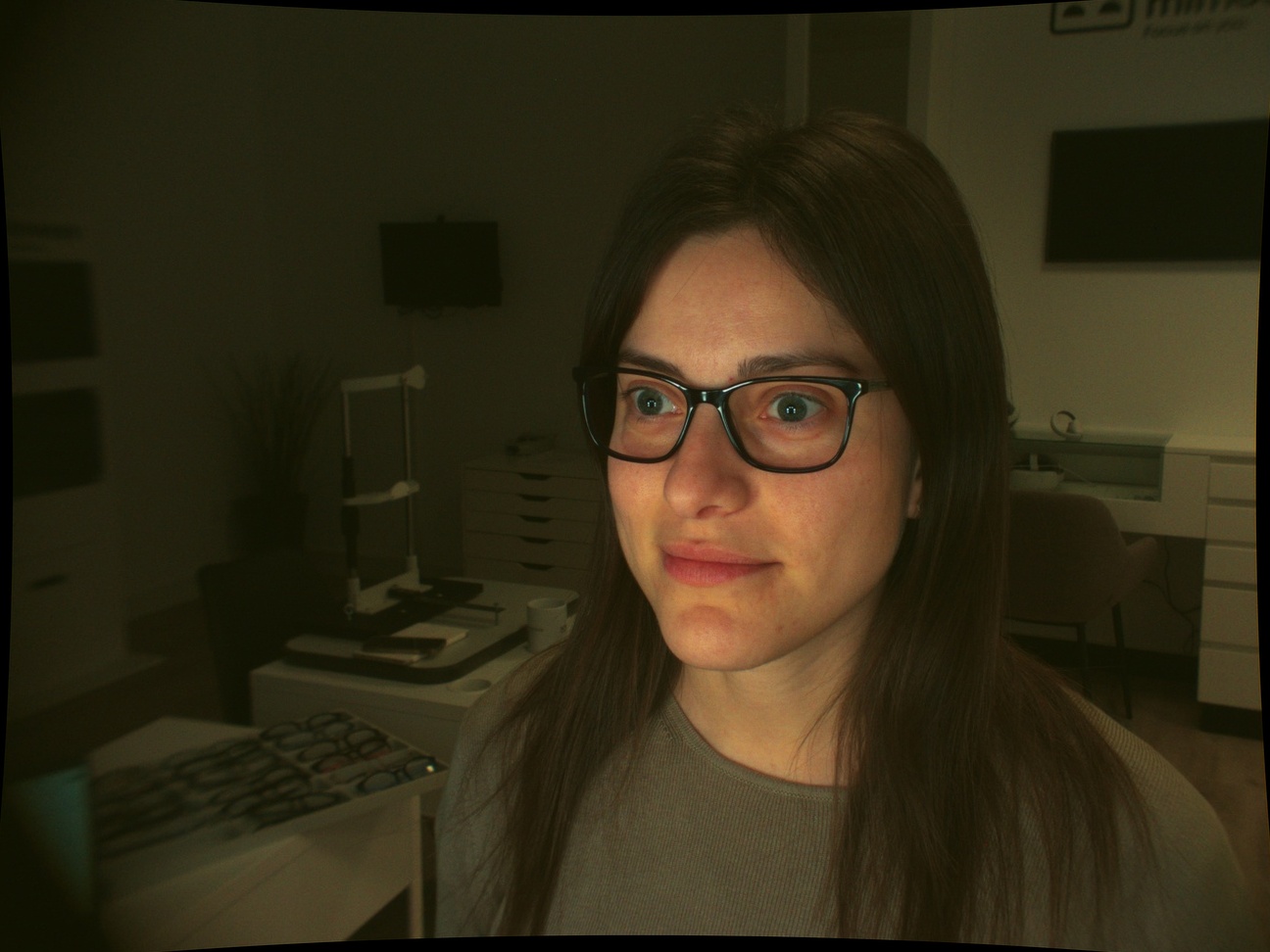}&
            \includegraphics[clip, angle=0,width=1.0\linewidth]{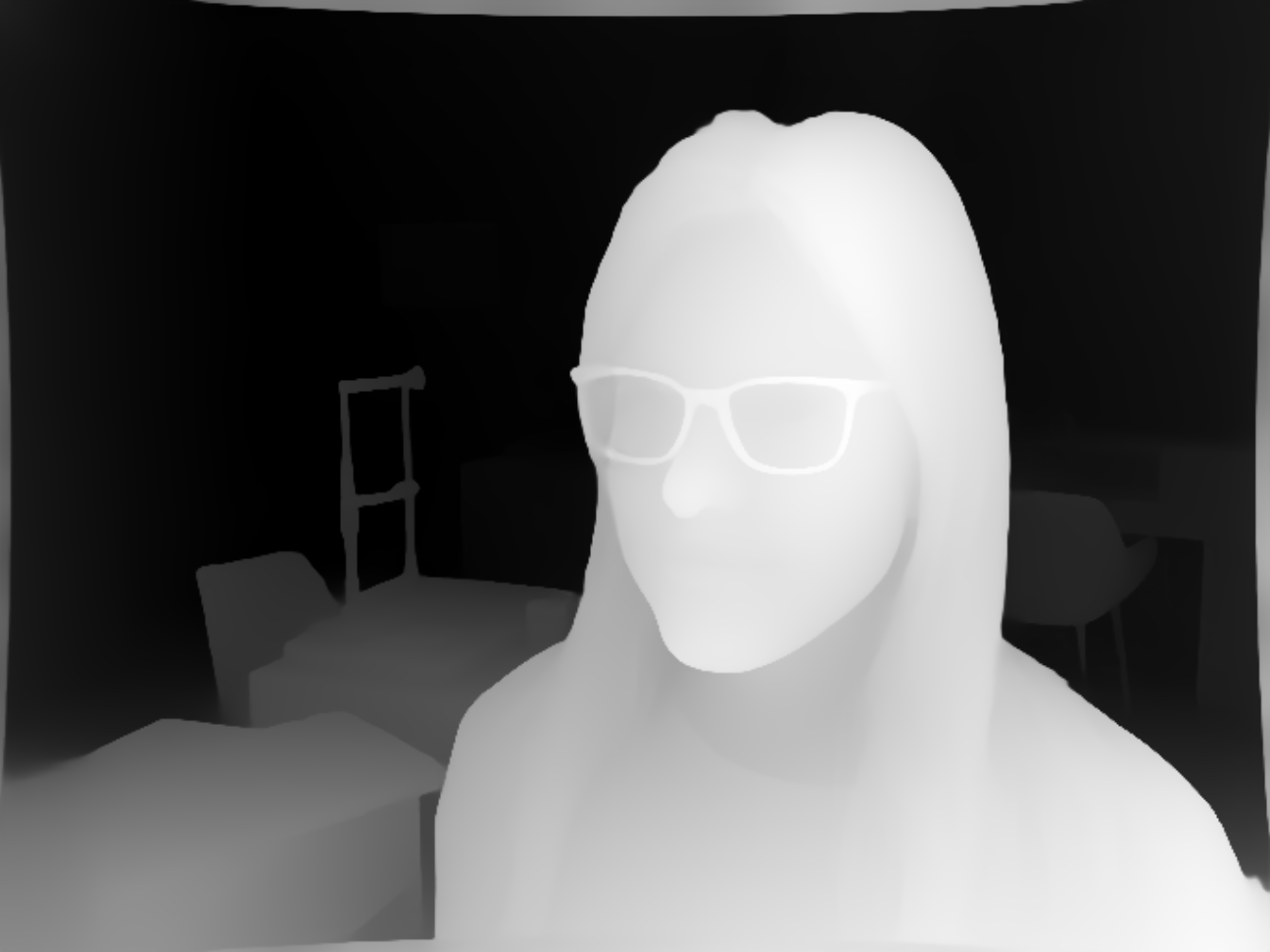}&
              \includegraphics[clip, angle=0,width=1.0\linewidth]{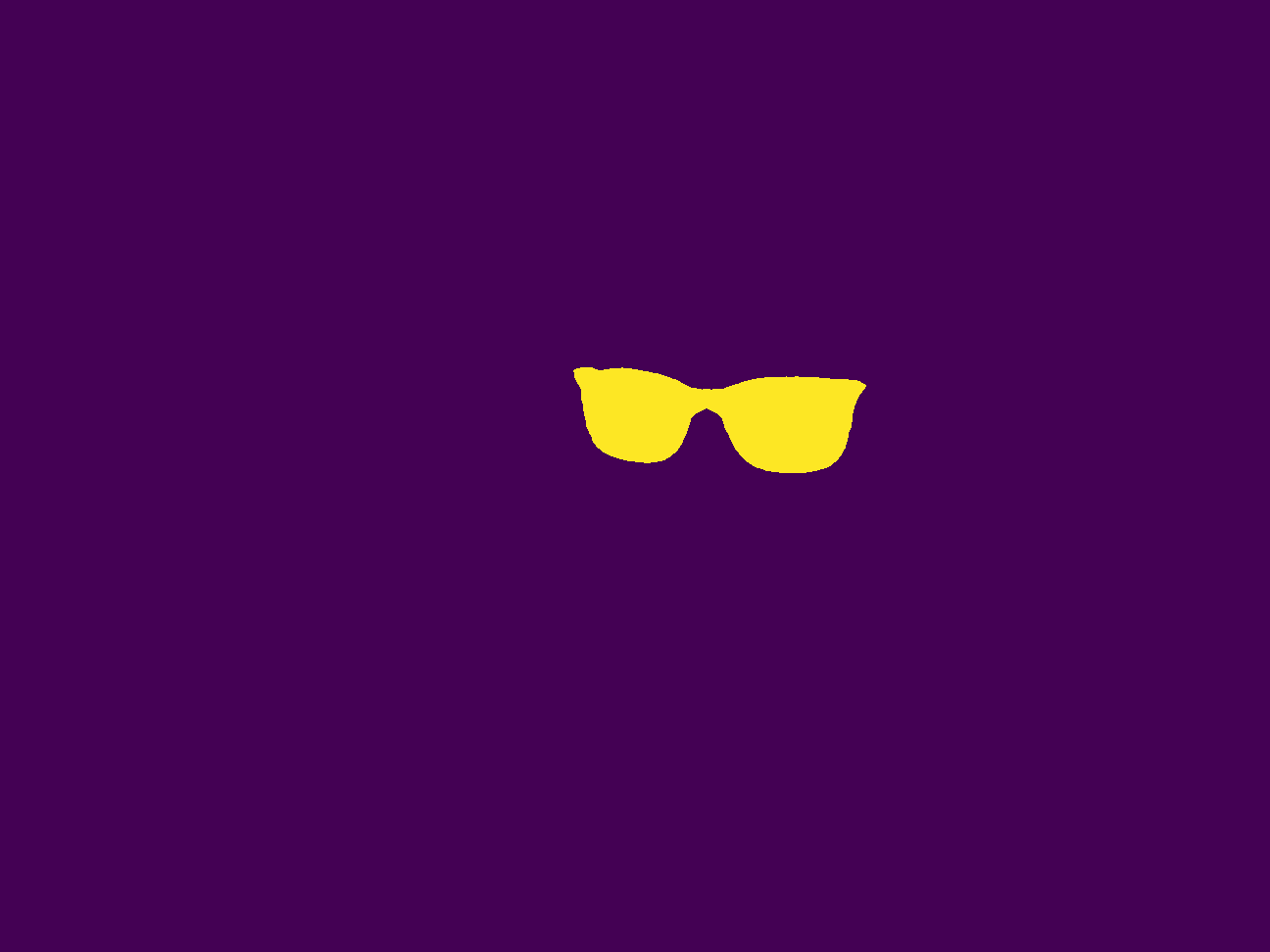}&
                \includegraphics[clip, angle=0,width=1.0\linewidth]{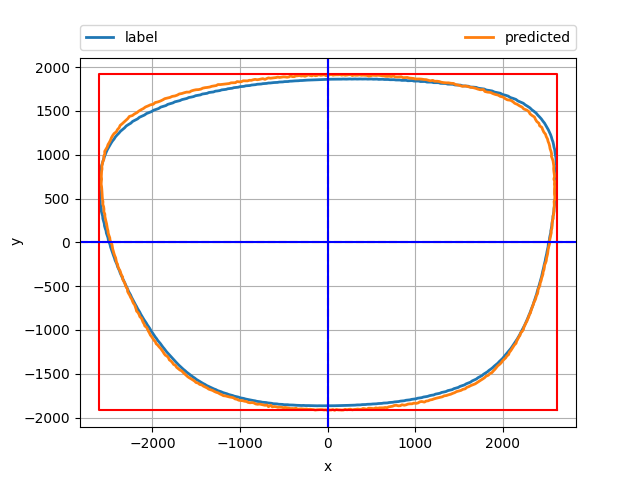}
          \end{tabular}}
    \caption{\textbf{Qualitative evaluation of our approach.} In all rows, the same information is provided. From left to right: an arbitrary image acquisition, the predicted depth, the predicted segmentation mask and the final trace output plot trace. In this last plot, the scale is in hundredths of a millimeters. Ground truth and our prediction are represented in blue and orange, respectively. Best viewed in color.}
    \label{fig:final-qualitative-results}
\end{figure*}

Fig.~\ref{fig:final-qualitative-results} presents representative examples of our estimations compared against the ground truth for the best combination on Table~\ref{tab:experiment-mvdcnn2heads}. Specifically, we highlight three scenarios: the best-performing case (top row), the median case (middle row), and the worst-performing one (bottom row). As shown, the poorest results arise in regions that are heavily occluded from multiple viewpoints. This effect is particularly evident in the upper corners of the eyeframe, where occlusions lead to increased prediction instability. Additionally, we observe that samples resembling the first two rows are more prevalent in the dataset than those similar to the third row. This observation supports the hypothesis that increasing the number and diversity of samples, particularly those covering a wider range of challenging conditions, would further improve the model’s overall performance.

\section{Conclusion}
This work introduces a novel computer vision–based framework for eyeframe measurement that overcomes key limitations associated with traditional mechanical tracing systems. The proposed approach leverages multi-view RGB imagery acquired from an InVision system, thereby removing the need for mechanical instrumentation, patron projection, and labor-intensive calibration procedures. The pipeline comprises three principal components: frame segmentation using a fine-tuned, SAM2-inspired model to isolate the eyeframe from the background; depth estimation via an off-the-shelf monocular method to recover 3D spatial information; and a multi-view reconstruction architecture to derive precise frame contour measurements. To support that, we introduce two dedicated datasets: one with high-quality mask annotations for segmentation, and another with ground-truth trace measurements for quantitative evaluation. We assess a range of model configurations, including variations in network size, input modalities, and multi-view fusion strategies, demonstrating that the proposed method achieves accurate and robust performance on real-world data while significantly reducing workflow complexity for opticians. The experimental results validate the feasibility of vision-based measurement systems within optical manufacturing and establish a promising foundation for attaining industry-standard, sub-millimeter precision.

\bibliographystyle{IEEEtran}
\bibliography{references}

\end{document}